\documentclass{article}

\usepackage{microtype}
\usepackage{graphicx}
\usepackage{subcaption}
\usepackage{booktabs}
\usepackage{hyperref}

\usepackage[preprint]{icml2026}  
\usepackage{float}
\usepackage{amsmath}
\usepackage{amssymb}
\usepackage{mathtools}
\usepackage{amsthm}
\usepackage{wrapfig}

\usepackage{booktabs}
\usepackage{multirow}
\usepackage{colortbl}
\usepackage[table]{xcolor}
\usepackage{makecell}
\usepackage{enumitem}
\usepackage[normalem]{ulem}


\definecolor{trainhdr}{RGB}{62,124,188}   
\definecolor{testhdr}{RGB}{230,126,34}    
\definecolor{hdrbg}{RGB}{245,247,250}     
\definecolor{oursbg}{RGB}{236,245,255}    
\definecolor{ablbg}{RGB}{250,250,250}     
\usepackage{soul}

\usepackage{soul, xcolor}

\usepackage{booktabs}
\usepackage{amssymb}
\usepackage{pifont}

\usepackage{placeins}
\usepackage{rotating}

\icmltitlerunning{HoRD: Robust Humanoid Control via \underline{H}istory-C\underline{o}nditioned \underline{R}einforcement Learning and Online \underline{D}istillation}

\begin{document}

\twocolumn[
  \icmltitle{HoRD: Robust Humanoid Control via \underline{H}istory-C\underline{o}nditioned \underline{R}einforcement Learning and Online \underline{D}istillation}

\begin{icmlauthorlist}
  \icmlauthor{Puyue Wang$^{*}$}{auck}
  \icmlauthor{Jiawei Hu$^{*}$}{unsw}
  \icmlauthor{Yan Gao}{cam}
  \icmlauthor{Junyan Wang}{adel}
  \icmlauthor{Yu Zhang}{mq}\\
  \icmlauthor{Gillian Dobbie}{auck}
  \icmlauthor{Tao Gu}{mq}
  \icmlauthor{Wafa Johal}{unimelb}
  \icmlauthor{Ting Dang}{unimelb}
  \icmlauthor{Hong Jia}{auck}
  \vskip 0.05in

{\small \href{https://tonywang-0517.github.io/hord/}{\raisebox{-2.0pt}{\includegraphics[height=10pt]{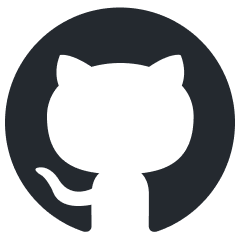}}
\textcolor{magenta}{\texttt{tonywang-0517.github.io/hord/}}}
}
\end{icmlauthorlist}

\icmlaffiliation{auck}{University of Auckland}
\icmlaffiliation{unsw}{University of New South Wales}
\icmlaffiliation{cam}{University of Cambridge}
\icmlaffiliation{adel}{University of Adelaide}
\icmlaffiliation{mq}{Macquarie University}
\icmlaffiliation{unimelb}{University of Melbourne}

\icmlcorrespondingauthor{Hong Jia}{hong.jia@auckland.ac.nz}

  \icmlkeywords{Robotics, Humanoid Control, Masked Imitation, Domain Randomization}

  \vskip 0.3in
]

\printAffiliationsAndNotice{\icmlEqualContribution}  


\begin{abstract}

Humanoid robots can suffer significant performance drops under small changes in dynamics, task specifications, or environment setup. We propose HoRD, a two-stage learning framework for robust humanoid control under domain shift. First, we train a high-performance teacher policy via history-conditioned reinforcement learning, where the policy infers latent dynamics context from recent state--action trajectories to adapt online to diverse randomized dynamics. Second, we perform online distillation to transfer the teacher's robust control capabilities into a transformer-based student policy that operates on sparse root-relative 3D joint keypoint trajectories. By combining history-conditioned adaptation with online distillation, HoRD enables a single policy to adapt zero-shot to unseen domains without per-domain retraining. Extensive experiments show HoRD outperforms strong baselines in robustness and transfer, especially under unseen domains and external perturbations. 
\end{abstract}

\section{Introduction}

Humanoid robots promise to unify locomotion, manipulation, and whole-body interaction in dynamic, human-centric environments. Recent advances in high-level planning and abstraction, ranging from vision–language–action (VLA) systems~\cite{ahn2022saycan,driess2023palme,brohan2023rt2,zhao2025cot,kim2024openvla} to learning-based planning and latent behavior modeling in reinforcement learning~\cite{janner2022planning,ghosh2023reinforcement}, have significantly improved semantic reasoning and task specification. Such high-level specifications are often formalized as goal-conditioned or multi-task reinforcement learning (RL) problems~\cite{schaul2015universal}. However, learning policies that map these high-level commands to stable, torque-level humanoid control remains a fundamental challenge, particularly under distribution shift where test-time dynamics, contact properties, or physics engines differ from training conditions.

Existing physics-based control policies are typically trained on dense motion-capture datasets~\cite{peng2018deepmimic,mahmood2019amass} within a single simulator with fixed dynamics parameters. These methods suffer from two critical limitations. First, the training distribution is limited and sparse: motion datasets cover only a narrow range of skeleton definitions, frame rates, and input modalities, leading to incomplete motion-manifold coverage and hindering scalable learning and fair method comparison. 
Second, policies overfit to source-domain dynamics and exhibit catastrophic failure under domain shift~\cite{cobbe2020leveraging}, such as when operating with different motion patterns, on different terrains, or when evaluated in different simulation engines. Recent masked imitation method (i.e., MaskedMimic)~\cite{tessler2024maskedmimic} addresses data fragmentation by reformulating control as masked motion inpainting, achieving visually plausible motions but lacking generalization robustness. When evaluated in a different physics engine, these policies fail to maintain stability, limiting their utility as reliable execution layers. 

Together, data fragmentation and domain shift expose a key gap: current methods lack a unified, robust representation from high-level motion specifications to torque-level control that generalizes across data sources and physical environments. This motivates a learning framework that (i) is able to track human-like motions from sparse motion commands, (ii) generalizes robustly under distribution shifts in physical dynamics or simulators, and (iii) is capable of \emph{adapting online} to unobserved dynamic changes (e.g., mass distribution, friction, actuator delay); otherwise, small model mismatches can compound into large tracking errors and destabilize torque control over long horizons.

We propose HoRD, a learning framework that maps sparse motion commands (formalized as root-relative 3D joint keypoint trajectories) to torque-level humanoid control with strong cross-domain generalization. HoRD uses a two-stage pipeline in which we (i) train a high-performance teacher torque policy via RL on large-scale real world human motion data under diverse randomized dynamics, and (ii) distill it into a transformer-based student that executes sparse keypoint commands.
Both teacher and student are conditioned by HCDR, a history-based module that infers a latent dynamics context from recent state--action trajectories for online compensation to handle domain shift, while the student additionally operates in SSJR, a standardized sparse-joint command representation that addresses data fragmentation by unifying fragmented datasets to process sparse keypoint commands from diverse motion sources.

This work makes the following contributions:
\begin{itemize}[leftmargin=*, itemsep=0.2em, topsep=0.2em, parsep=0pt, partopsep=0pt]
    \item We propose a two-stage learning framework that maps sparse root-relative 3D joint keypoint trajectories to torque-level humanoid control with strong cross-domain generalization.
    \item We introduce a history-conditioned RL framework that infers latent dynamics parameters from recent state--action sequences, enabling online adaptation to distribution shift and significantly improving cross-domain generalization and zero-shot transfer. We also released a large-scale trajectory dataset of 7,000+ diverse humanoid motions (100+ hours), to help the community train and evaluate their specific humanoid robots.


    \item Extensive evaluations show that HoRD consistently outperforms state-of-the-art baselines, achieving up to 14.2\% higher success rates under domain shift and enabling reliable zero-shot transfer from IsaacLab to an unseen physics engine (Genesis), where prior methods degrade or fail catastrophically.
\end{itemize}
\section{Related Work}
\label{sec: related}

\paragraph{Physics-guided policy learning for humanoid control.}
Physics-based character control learns humanoid behaviors by optimizing torque-level policies under contact-rich dynamics in simulation. Early work focused on imitation and motion tracking of individual reference clips, producing realistic locomotion and agile behaviors from curated demonstrations~\cite{peng2017deeploco,peng2018deepmimic}. More recently, the increasing availability of large-scale motion capture datasets such as AMASS~\cite{mahmood2019amass} has enabled methods to scale toward training over diverse and heterogeneous motion collections. Such policy learning across broad skill distributions has been extensively studied in the context of multi-task optimization and transfer, with challenges including gradient interference and negative transfer across behaviors~\cite{sener2018multi,yu2020pcgrad,wang2025soft, zhang2024proactive}.
Along this scaling direction, some work has explored two complementary paradigms to support large-scale motion learning: expert-based decomposition with mixture-of-experts routing or distillation frameworks~\cite{won2020scalable,shazeer2017outrageously,zhou2022expertchoice,fedus2022switch}, such as Hover~\cite{he2025hover} and ExBody2~\cite{ji2024exbody2}; and unified-policy training strategies that acquire broad motion priors through adversarial imitation or masked objectives without explicit expert routing, as demonstrated in AMP~\cite{peng2021amp} and OmniH2O~\cite{he2025omnih2o}.
However, a fundamental limitation persists: existing methods require dense reference trajectories (i.e., full joint-space motion target) during both training and deployment. This creates a critical deployment gap, as real-world motion planners, video-to-motion systems, and high-level controllers typically provide only \emph{sparse keypoint commands} (e.g., only the 3D positions of key body joints such as shoulders, hips, and knees extracted from video, rather than complete joint-space trajectories with all 20+ joint angles) rather than complete joint trajectories. Moreover, heterogeneous motion sources often use incompatible skeleton definitions and temporal resolutions, further complicating integration.

HoRD addresses this limitation through SSJR, a standardized sparse-joint representation that unifies fragmented motion sources into a single training distribution, enabling policies to operate directly from sparse commands. By adopting a teacher--student framework, HoRD trains a robust and unified expert policy using dense references (which are available in simulation), then distills this expertise into a lightweight student policy that executes from sparse SSJR commands, bridging the gap between simulation training and real-world deployment.

\begin{figure*}[t]
    \centering
    \includegraphics[width=\linewidth]{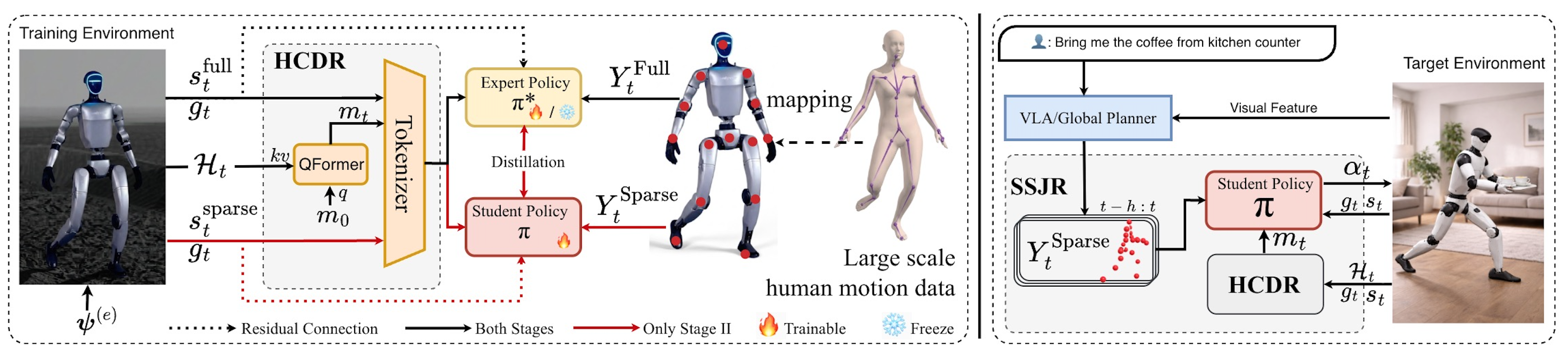}
    \caption{\textbf{Framework overview.} Two-stage teacher--student learning pipeline for robust humanoid control under partial observability. \textbf{Stage I:} an expert policy $\pi^\star$ is trained with PPO in simulation using privileged full-state observations $\mathbf{s}_t^{\text{full}}$, dense future motion intent $\mathbf{Y}_t^{\text{full}}$, and episode-level domain randomization $\boldsymbol{\psi}^{(e)}$. A shared \textsc{HCDR} module encodes the interaction history $\mathcal{H}_t$ into a temporal memory embedding $\mathbf{m}_t$ for online dynamics inference and adaptive modulation. \textbf{Stage II:} a deployable student policy $\pi$ receives only sparse proprioception $\mathbf{s}_t^{\text{sparse}}$, environment context $\mathbf{g}_t$, and standardized motion commands $\mathbf{Y}_t^{\text{sparse}}$ via \textsc{SSJR}, and is trained by distillation to match the expert's actions. \textsc{SSJR} maps a global planner command into a platform-agnostic sparse-joint command interface, enabling cross-platform transfer while \textsc{HCDR} provides in-context adaptation to latent dynamics during deployment.}
    \label{fig:overview}
\vspace{-1em}
\end{figure*}

\paragraph{Cross-domain generalization and dynamics adaptation.}
Generalization across domains (e.g, different motions, terrains and environments) with varying dynamics is a longstanding challenge in reinforcement learning, particularly for contact-rich control problems~\cite{cobbe2019quantifying}.
A common strategy is domain randomization, which exposes policies to a distribution of physical parameters during training to enable zero-shot transfer~\cite{tobin2017domainrand,campanaro2024learning}, as demonstrated by recent humanoid controllers such as Hover~\cite{he2025hover}, ExBody2~\cite{ji2024exbody2}, and OmniH2o~\cite{he2025omnih2o}.
An alternative line of work focuses on target-domain adaptation, such as residual or delta-action learning\cite{he2025asap}, which leverages limited target-domain data to compensate for dynamics mismatch~\cite{chebotar2019closing}.
However, both approaches have limitations. Domain randomization alone is insufficient. Even with extensive parameter coverage, policies cannot anticipate all possible dynamic variations, and the learned representations often fail to adapt when encountering novel dynamics at test time. Recent humanoid control systems demonstrate strong performance under nominal training conditions, but exhibit noticeable degradation under distribution shifts, including changes in contact dynamics, actuator delays, or physics engines~\cite{he2025hover,he2025omnih2o,ji2024exbody2}. Moreover, aggressive randomization can make optimization more difficult and slow convergence for high-dimensional humanoid systems. Target-domain adaptation methods, while effective, require additional interaction and fine-tuning in the target domain, which is impractical for real-world deployment where collecting and labeling target-domain data may be costly or unsafe.
HoRD resolves this through HCDR, a novel history-conditioned dynamics adaptation mechanism that learns to infer latent dynamics parameters from recent state--action sequences.

Unlike prior adaptation methods that require target-domain data, HCDR enables \emph{online adaptation} at test time by encoding how the environment has recently responded to policy actions. This learned representation enables zero-shot transfer without any target-domain fine-tuning, addressing the fundamental limitation that static domain randomization cannot fully capture the dynamics variations encountered in deployment. Appendix~\ref{subsec:method_comparison} provides a structured comparison of representative prior humanoid control frameworks and our method along three deployment-relevant capabilities: unified skill coverage, sparse-command generalization, and explicit online dynamics adaptation

\section{Method}

We propose a framework for robust, anticipatory humanoid control that generalizes across domains. Our approach decomposes the control problem into three primary components: \emph{SSJR} for standardized motion representation, \emph{Episode-Level Domain Randomization} for distributional robustness, and \emph{HCDR} for history-conditioned dynamics representation.

\subsection{Problem Definition}
\label{subsec:problem}
We formalize the task of learning an adaptive, domain-agnostic control policy as a \emph{Contextual Partially Observable Markov Decision Process} (C-POMDP)~\cite{hallak2015contextual}. This formulation explicitly models both the robot's partial observability and latent variations in environment dynamics.

\paragraph{Latent Environment Dynamics.} 
Each episode \(e\) is associated with an unobserved, episode-specific physical parameter vector \(\boldsymbol{\psi}^{(e)} \sim P(\Psi)\) that encodes physical properties such as control delay, body mass scaling, and gravity magnitude. The robot state \(\mathbf{s}_t\) evolves according to parameterized transitions that depend on \(\boldsymbol{\psi}^{(e)}\), where \(\mathbf{s}_t\) represents the robot's configuration and dynamics state at time \(t\). The control modulation signal \(\boldsymbol{\alpha}_t\) allows the robot to adapt its behavior to the specific latent dynamics of the current episode.

\paragraph{Observations and Anticipatory Input.} 
To enable anticipatory control, the robot receives future motion intentions \(\mathbf{Y}_t\) over a short-horizon lookahead, which provide guidance for upcoming movements. The robot also maintains historical interactions \(\mathcal{H}_t\) (formally defined in \S\ref{subsec:framework}), capturing recent state-action pairs that encode how the environment has responded to previous control commands. Additionally, the robot observes environmental context \(\mathbf{g}_t\), such as terrain geometry or obstacles, which informs its control decisions.

\paragraph{Policy Objective.} 
The goal is to learn a policy \(\pi\) that generates adaptive proportional-derivative (PD) targets as the control modulation signal \(\boldsymbol{\alpha}_t\) based on both the observed history and future motion intent. The policy leverages a temporal memory embedding \(\mathbf{m}_t\) (see \S\ref{subsec:HCDR}) that captures latent dynamics information, enabling the policy to adapt its control strategy in real-time. The optimal policy maximizes performance across diverse physical and environmental conditions, where \(t\) denotes discrete time steps, \(T\) is the episode horizon (an episode starts from initialization and terminates when the tracking error exceeds a threshold), and \(\gamma \in [0,1)\) is the discount factor.

\subsection{Framework Overview}
\label{subsec:framework}
We propose a two-stage learning framework designed to bridge the gap between high-fidelity simulation and real-world deployment under partial observability as shown in Figure~\ref{fig:overview}. The framework leverages a teacher-student distillation paradigm: a full-observation expert policy provides supervision to a deployable student policy that operates under realistic, sparse sensory inputs.

Both the expert and student policies utilize a shared mechanism for processing historical interactions. They collect a history window
\begin{equation}
\mathcal{H}_t = \{ (\mathbf{s}_{t-k}, \boldsymbol{\alpha}_{t-k}) \}_{k=1}^{K},
\end{equation}
containing recent state-action pairs, where $K$ is the history window length. This history is processed by the proposed \textsc{HCDR} module (see \S\ref{subsec:HCDR}) to produce a temporal memory embedding $\mathbf{m}_t$ that captures latent dynamics information and serves as a key input to both policies, enabling adaptive control based on how the environment responds to previous actions.

\paragraph{Stage I: Full-Observation Expert Training.}  
In the first stage, we train an expert (teacher) policy $\pi^\star$ in simulation with privileged access to the full state $\mathbf{s}_t^{\text{full}}$, including link velocities, contact forces, and precise joint torques, as well as dense future motion intent $\mathbf{Y}_t^{\text{full}}$ derived from high-frequency motion capture data. The expert obtains temporal memory embedding $\mathbf{m}_t$ from historical interactions $\mathcal{H}_t$ (as mentioned above). The expert then predicts the control modulation signal
\begin{equation}
\boldsymbol{\alpha}_t^{(1)} = f_{\theta_1}(\mathbf{m}_t, \mathbf{s}_t^{\text{full}}, \mathbf{g}_t, \mathbf{Y}_t^{\text{full}}),
\end{equation}
where $\boldsymbol{\alpha}_t^{(1)}$ is the expert policy output, $f_{\theta_1}$ is the expert policy network parameterized by $\theta_1$, $\mathbf{s}_t^{\text{full}}$ is the full state observation, $\mathbf{g}_t$ encodes environmental context, and $\mathbf{Y}_t^{\text{full}}$ is the dense future motion intent.

The expert is trained using Proximal Policy Optimization (PPO) to maximize the expected discounted return across domain-randomized episodes:
\begin{equation}
J(\pi^\star) = \mathbb{E}_{\boldsymbol{\psi}^{(e)},\,\tau} \Bigg[ 
\sum_{t=0}^T \gamma^t \, r(\mathbf{s}_t, \boldsymbol{\alpha}_t^{(1)}, \mathbf{Y}_t^{\text{full}}) 
\Bigg],
\end{equation}
where $J(\pi^\star)$ is the expert policy objective function, $\pi^\star$ denotes the expert policy, $\tau = \{(\mathbf{s}_t, \boldsymbol{\alpha}_t^{(1)})\}_{t=0}^T$ is a trajectory generated by the expert policy under the latent dynamics $\boldsymbol{\psi}^{(e)}$, and $r(\cdot)$ is a reward function that penalizes deviations from desired motion, loss of balance, excessive control effort, and falls. By training over a diverse distribution of $\boldsymbol{\psi}^{(e)}$ using domain randomization (see \S\ref{subsec:dr}), the expert policy learns to handle a wide range of latent dynamics and environment variations.

\paragraph{Stage II: Sparse-Observation Student.}  
The second stage distills the expert's behavior into a student policy $\pi_{\theta_2}$ that operates under realistic sensory constraints. The student receives sparse observations $\mathbf{s}_t^{\text{sparse}} \subset \mathbf{s}_t^{\text{full}}$ and sparse future motion cues $\mathbf{Y}_t^{\text{sparse}}$, but uses the same temporal memory embedding $\mathbf{m}_t$ (obtained from $\mathcal{H}_t$ as mentioned above) and the proposed \textsc{SSJR} representation (see \S\ref{subsec:SSJR}) to leverage historical interactions and standardized motion commands.

The student outputs the control modulation
\begin{equation}
\boldsymbol{\alpha}_t^{(2)} = f_{\theta_2}(\mathbf{m}_t, \mathbf{s}_t^{\text{sparse}}, \mathbf{g}_t, \mathbf{Y}_t^{\text{sparse}}),
\end{equation}
where $\boldsymbol{\alpha}_t^{(2)}$ is the student policy output, $f_{\theta_2}$ is the student policy network parameterized by $\theta_2$, $\mathbf{s}_t^{\text{sparse}}$ is the sparse state observation, and $\mathbf{Y}_t^{\text{sparse}}$ is the sparse future motion intent. The student is trained via supervised distillation to match the expert:
\begin{equation}
\label{eq:distill}
\mathcal{L}_{\text{distill}}(\theta_2) = 
\mathbb{E}_{\boldsymbol{\psi}^{(e)},\,\tau} \Bigg[ 
\sum_{t=1}^{T} \ell\big(\boldsymbol{\alpha}_t^{(2)}, \boldsymbol{\alpha}_t^{(1)}\big) 
\Bigg],
\end{equation}
where $\mathcal{L}_{\text{distill}}$ is the distillation loss, and $\ell(\cdot)$ is the mean squared error (MSE).

By sharing the temporal memory representation $\mathbf{m}_t$ and the SSJR motion interface across both stages, the student learns to recover the expert's anticipatory behavior using only its own historical interactions and partial observations, resulting in a robust and domain-agnostic policy that enables reliable control across a wide range of physical and environmental conditions.

\subsection{SSJR: Standardized Sparse-Joint Representation}
\label{subsec:SSJR}
To decouple high-level motion specification from low-level torque control for humanoid robots, we introduce a standardized sparse-joint representation (SSJR). SSJR enables leveraging large-scale human motion data to train robot controllers without requiring platform-specific robots torque demonstrations, thereby improving scalability and cross-platform generalization.

\paragraph{Motion Abstraction.}  
SSJR represents future motion intentions $\mathbf{Y}_t$ using a sparse set of key joints $\mathcal{J}$ defined in a canonical human skeletal model (e.g., SMPL-X). The motion commands over a short-horizon lookahead $H$ are expressed as:
\begin{equation}
\mathbf{Y}_t = \big\{ \mathbf{q}_{j, t+h}^{\text{cmd}} \big\}_{j \in \mathcal{J}, \, h=1}^{H},
\end{equation}
where $\mathbf{q}^{\text{cmd}}$ denotes the relative joint configurations. By projecting human motion data from large-scale motion capture datasets (e.g., AMASS) onto this canonical sparse-joint representation, SSJR abstracts away platform-specific details while preserving task-relevant motion information.

\paragraph{Policy Integration.}  
At execution, the student policy $\pi_{\theta_2}$ receives the SSJR commands $\mathbf{Y}_t$ along with robot proprioception and environmental context, and outputs torque modulation signals $\boldsymbol{\alpha}_t$. By standardizing the input across human datasets and humanoid platforms, SSJR ensures that the learned policy generalizes robustly to diverse robots and environmental conditions.

SSJR offers several advantages. \emph{Data scalability}: The sparse-joint representation allows the use of widely available human motion datasets without costly torque supervision.  
\emph{Cross-platform transfer}: By abstracting away robot-specific actuation details, SSJR supports generalization across heterogeneous humanoid morphologies.  
\emph{Anticipatory control}: Coupled with temporal memory (\textsc{HCDR}, \textit{cf}. \S\ref{subsec:HCDR}) and domain randomization, SSJR enables the policy to interpret future motion intentions and adapt its torque commands accordingly.

\subsection{Episode-Level Domain Randomization}
\label{subsec:dr}
To promote robust cross-domain generalization and prevent overfitting to simulator-specific artifacts, we adopt an \emph{Episode-Level Domain Randomization} strategy. Rather than applying uncorrelated noise at each timestep, we sample $\boldsymbol{\psi}^{(e)}$ (as defined in \S\ref{subsec:problem}) at the start of each episode $e$ and hold it fixed for the duration of the episode. By maintaining temporally consistent dynamics shifts, the policy is encouraged to infer latent physical properties from its observations, rather than memorizing short-term fluctuations.

The robot state evolves according to a family of parameterized transitions:
\begin{equation}
\mathbf{s}_{t+1} = f(\mathbf{s}_t, \boldsymbol{\alpha}_t; \boldsymbol{\psi}^{(e)}) + \boldsymbol{\omega}_t,
\end{equation}
where $\mathbf{s}_t$ is the state at time $t$, $\boldsymbol{\alpha}_t$ is a control modulation applied to the low-level actuators, and $\boldsymbol{\omega}_t \sim \mathcal{N}(0, \Sigma_\omega)$ captures unmodeled disturbances.

We randomize $\boldsymbol{\psi}^{(e)}$ along four key dimensions: (1) inertial properties, including link masses, center-of-mass locations, and joint damping coefficients; (2) contact dynamics, such as ground friction coefficients $\mu$ and contact stiffness; (3) probabilistic system latency delays $\delta \in \{1, 2\}$ in executing control commands to emulate communication and actuator delays; and (4) periodic random external forces applied to the robot's torso to simulate environmental disturbances and contact irregularities.

By sampling coherent parameter vectors per episode, the agent experiences dynamics that are both diverse and temporally consistent. This exposes the policy to distributional shifts similar to those encountered when transferring between different physics engines (e.g., IsaacLab vs. Genesis) or during real-world deployment, enabling robust and adaptive control under previously unseen conditions.

\subsection{HCDR: History-Conditioned Dynamics Representation}
\label{subsec:HCDR}
To adapt to latent environment parameters $\boldsymbol{\psi}^{(e)}$ (defined in \S\ref{subsec:problem}) in real-time, we introduce \textsc{HCDR}, a latent representation module that performs online dynamics inference from historical interactions.

\paragraph{Latent Dynamics Inference.}  
\textsc{HCDR} models historical interactions as evidence for latent dynamics using a Query-Transformer (Q-Former) architecture, rather than flattening them into a simple vector. We define a set of learnable latent tokens $\mathbf{m}_0 \in \mathbb{R}^{d_m \times N}$ that encode a shared prior over dynamics configurations encountered during training, where $d_m$ is the embedding dimension and $N$ is the number of tokens. The temporal memory embedding $\mathbf{m}_t$ is computed via cross-attention between the prior tokens and the history $\mathcal{H}_t$ (defined in \S\ref{subsec:framework}):
\begin{equation}
\mathbf{m}_t = \mathrm{QFormer}(\mathbf{m}_0, \mathcal{H}_t),
\end{equation}
where $\mathbf{m}_t \in \mathbb{R}^{d_m}$ is the temporal memory embedding at time $t$.

This mechanism allows the module to selectively extract history components that are most informative for identifying the current latent dynamics regime.

\paragraph{Adaptive Modulation.}  
The resulting embedding $\mathbf{m}_t$ acts as a compact \emph{dynamics fingerprint}, capturing unobserved physical properties such as surface friction, unexpected mass distribution, or actuator delays. By conditioning the control modulation $\boldsymbol{\alpha}_t$ on $\mathbf{m}_t$, the policy can adapt its behavior ``in context'', adjusting gait, balance, and torque outputs based on how the environment responds to previous actions. In essence, \textsc{HCDR} enables online system identification and adaptive control without direct access to the latent parameters $\boldsymbol{\psi}^{(e)}$.

\vspace{-0.5em}
\section{Experiments}
\label{sec:experiments}

We evaluate HoRD on large-scale AMASS-based benchmarks and a suite of robustness scenarios. We design a transformer-based architecture with a 6-layer encoder using 8 attention heads. The actor head uses a 4-layer MLP and outputs the mean actions. Detailed model structures and RL settings are discussed in appendix~\ref{app:training} and ~\ref{app:model}.

\begin{table*}[t]
\centering
\caption{\textbf{Overall performance} of HoRD on the test set across 4 environments.
Genesis results are evaluated in a zero-shot manner without collecting test-domain data for retraining.
Best results are bold; second-best are underlined.}
\small
\setlength{\tabcolsep}{4.8pt}
\renewcommand{\arraystretch}{1.1}
\begin{tabular}{lccc|ccc|ccc|ccc}
\toprule
& \multicolumn{3}{c|}{\textcolor{trainhdr}{\textbf{IsaacLab (ID)}}}
& \multicolumn{3}{c|}{\textcolor{trainhdr}{\textbf{IsaacLab + DR (ID)}}}
& \multicolumn{3}{c|}{\textcolor{testhdr}{\textbf{Genesis (OOD)}}}
& \multicolumn{3}{c}{\textcolor{testhdr}{\textbf{Genesis + DR (OOD)}}} \\
\cmidrule(lr){2-4} \cmidrule(lr){5-7} \cmidrule(lr){8-10} \cmidrule(lr){11-13}

\textbf{Method}
& \textbf{Succ.} & $\mathbf{E_{\text{g-mpjpe}}}$ & $\mathbf{E_{\text{mpjpe}}}$
& \textbf{Succ.} & $\mathbf{E_{\text{g-mpjpe}}}$ & $\mathbf{E_{\text{mpjpe}}}$
& \textbf{Succ.} & $\mathbf{E_{\text{g-mpjpe}}}$ & $\mathbf{E_{\text{mpjpe}}}$
& \textbf{Succ.} & $\mathbf{E_{\text{g-mpjpe}}}$ & $\mathbf{E_{\text{mpjpe}}}$ \\
& (\%)$\uparrow$ & (mm)$\downarrow$ & (mm)$\downarrow$
& (\%)$\uparrow$ & (mm)$\downarrow$ & (mm)$\downarrow$
& (\%)$\uparrow$ & (mm)$\downarrow$ & (mm)$\downarrow$
& (\%)$\uparrow$ & (mm)$\downarrow$ & (mm)$\downarrow$ \\
\midrule

MaskedMimic
& 32.1 & 376 & 182
& $<$10 & 713 & 283
& $<$10 & 742 & 322
& $<$10 & 738 & 326 \\

OmniH2O
& 85.2 & 266 & 132
& 83.2 & 282 & 168
& 72.3 & 312 & 165
& \underline{70.2} & \underline{335} & \underline{191} \\

ExBody2
& \underline{86.6} & \underline{247} & \underline{108}
& \underline{85.9} & \underline{237} & \underline{128}
& \underline{73.1} & \underline{305} & \underline{157}
& 69.4 & 342 & 202 \\

Hover
& 71.2 & 278 & 138
& 67.9 & 375 & 196
& 16.2 & 722 & 258
& 15.5 & 746 & 282 \\

\midrule
\rowcolor{blue!6}
\textbf{HoRD (Ours)}
& \textbf{90.7} & \textbf{102} & \textbf{76}
& \textbf{88.4} & \textbf{124} & \textbf{87}
& \textbf{86.0} & \textbf{162} & \textbf{96}
& \textbf{84.4} & \textbf{171} & \textbf{108} \\

\bottomrule
\end{tabular}
\label{tab:overall_performance}
\vspace{-0.5em}
\end{table*}

\subsection{Experimental Setup}
\label{subsec:setup}

\textit{Robot and environment.} All experiments are conducted on a Unitree-style G1 humanoid with 29 actuated DoF, controlled at the PD level at 50\,Hz. Training is performed in Isaac Lab~\cite{mittal2025isaac} with contact-rich rigid-body dynamics, while evaluation additionally includes zero-shot transfer to the Genesis physics engine~\cite{zhou2024genesis}.
Episodes follow the duration of each target motion clip and terminate early if the tracking error diverges or the robot falls. Detailed environment and deployment specifications, together with the full reward formulation and stability constraints, are provided in Appendix~\ref{app:environment} and Appendix~\ref{subsec:reward_design}.

\textit{Motion dataset and commands.} We use the AMASS motion corpus~\cite{mahmood2019amass}, covering locomotion, balance, manipulation-style poses, and recovery motions.
Motion clips are segmented, retargeted to the humanoid, and converted into sparse key-joint trajectories using the proposed SSJR representation. Details of motion retargeting, filtering, SSJR and teacher/student observation specifications and input dimensions, are provided in Appendix~\ref{subsec:SSJR_example},~\ref{subsec:rl_env}, and~\ref{subsec:dataset_processing}.

\textit{Baselines.} We compare HoRD against strong physics-based and masked-imitation baselines, including MaskedMimic~\cite{tessler2024maskedmimic}, OmniH2O~\cite{he2025omnih2o}, ExBody2~\cite{ji2024exbody2}, and Hover~\cite{he2025hover}.
All methods are trained under the same environment configuration. Detailed baseline configurations and implementation notes are provided in Appendix~\ref{subsec:baseline_details}.

\textit{Evaluation metrics.}
In evaluation, we report:
(1) \emph{success rate} (Succ.\,), defined as the percentage of episodes that complete without falling,
(2) global pose error $\mathbf{E_{\text{g-mpjpe}}}$, and
(3) local pose error $\mathbf{E_{\text{mpjpe}}}$.
All results are averaged across multiple random seeds. Full experimental details, including environment configuration, metric definitions, and PPO training hyperparameters, in Appendix~\ref{app:environment} and Appendix Table~\ref{tab:hyperparams}.

\subsection{In-Domain Performance}

Table~\ref{tab:overall_performance} reports the test performance of HoRD and representative baselines on a test set of around 800 motions from AMASS. All methods are trained in IsaacLab using identical training data, while evaluation is conducted under four settings: IsaacLab and Genesis, each with and without additional domain randomization (DR). Genesis represents an unseen physics engine, and no test-domain data are collected for retraining (zero-shot).

Across all environments (w/o and w DR) and metrics, HoRD consistently achieves the best performance. In the in-domain environment (IsaacLab), HoRD attains a $90.7\%$ success rate with substantially lower motion tracking errors, reducing $\mathbf{E_{\text{g-mpjpe}}}$ and $\mathbf{E_{\text{mpjpe}}}$ by more than $2\times$ compared to strong physics-based baselines. This advantage persists under IsaacLab with DR, indicating that HoRD maintains stable execution even when additional observation noise and perturbations are introduced at test time.

In contrast, MaskedMimic performs poorly even in the seen environment, achieving only $32.1\%$ success and failing almost entirely ($<10\%$) under domain randomization. This behavior is consistent with its design focus on animation-quality motion synthesis rather than torque-level execution on a complex humanoid platform such as Unitree~G1. While other baselines achieve reasonable performance in the training environment, their success rates and tracking accuracy degrade notably once domain shifts are introduced.

\subsection{Cross-Domain Generalization}

We further examine cross-domain generalization (w/o and w DR) by evaluating all methods in Genesis (Table~\ref{tab:overall_performance} right), an unseen environment with different contact dynamics and numerical properties. (Detailed domain-shift factors are provided in Appendix~\ref{subsec:dr}). This setting tests whether a controller can generalize beyond the simulator used during training. As shown in Table~\ref{tab:overall_performance}, most baselines experience substantial performance degradation when transferred from IsaacLab to Genesis, even without additional domain randomization. Specifically, OmniH2O, ExBody2, and Hover suffer large drops in success rate and tracking accuracy, suggesting limited sim-to-sim robustness. MaskedMimic fails entirely in Genesis, reflecting its inability for sim-to-sim transfer.

\begin{figure*}[t]
    \centering
    \includegraphics[width=0.95\textwidth]{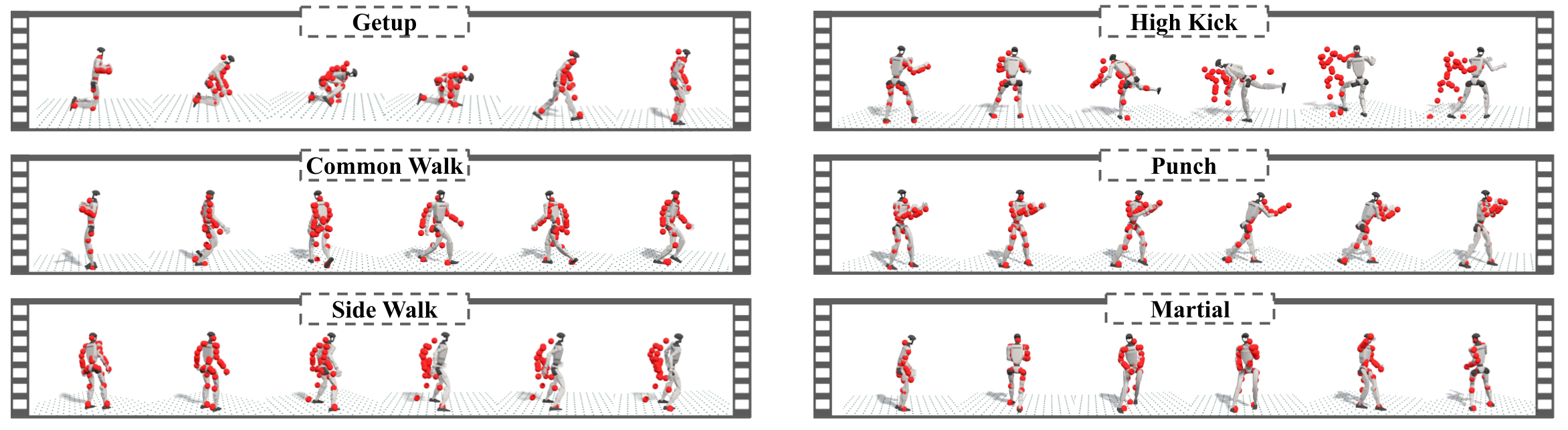}
    \vspace{-0.8em}
    \caption{Results of HoRD on six representative motions, while red markers indicate ground-truth skeleton joints. Qualitative comparison results with baselines are shown in Appendix Fig. 5. With DR and HCDR, the robot is able to mimic a range of different human actions.
}
    \label{fig:motion}
\end{figure*}

In contrast, HoRD exhibits strong zero-shot transfer to Genesis, achieving an $86.0\%$ success rate without retraining. When additional domain randomization is applied at test time, HoRD retains high performance ($84.4\%$ success), while all baselines collapse to near-failure regimes. We attribute this robustness to two factors. First, HoRD is trained under extensive domain randomization that covers a broad range of physical parameters. Second, the student policy additionally receives mild observation noise during training, encouraging adaptation to imperfect and shifted sensing conditions. Together with HCDR, which conditions control on recent state-action histories, HoRD is able to infer latent dynamics discrepancies online and adjust its execution accordingly.

\subsection{Representative Motion Case Studies}
\label{subsec:motions}

To better understand how HoRD behaves across diverse motion patterns beyond aggregate metrics, we conduct a motion-level analysis on six representative motions: common walk, side walk, get-up, punch, high kick, and martial. These motions vary substantially in contact patterns, balance requirements, and action aggressiveness.

\begin{figure}[t]
    \centering
    \begin{subfigure}[t]{0.49\linewidth}
        \centering
        \includegraphics[width=\linewidth]{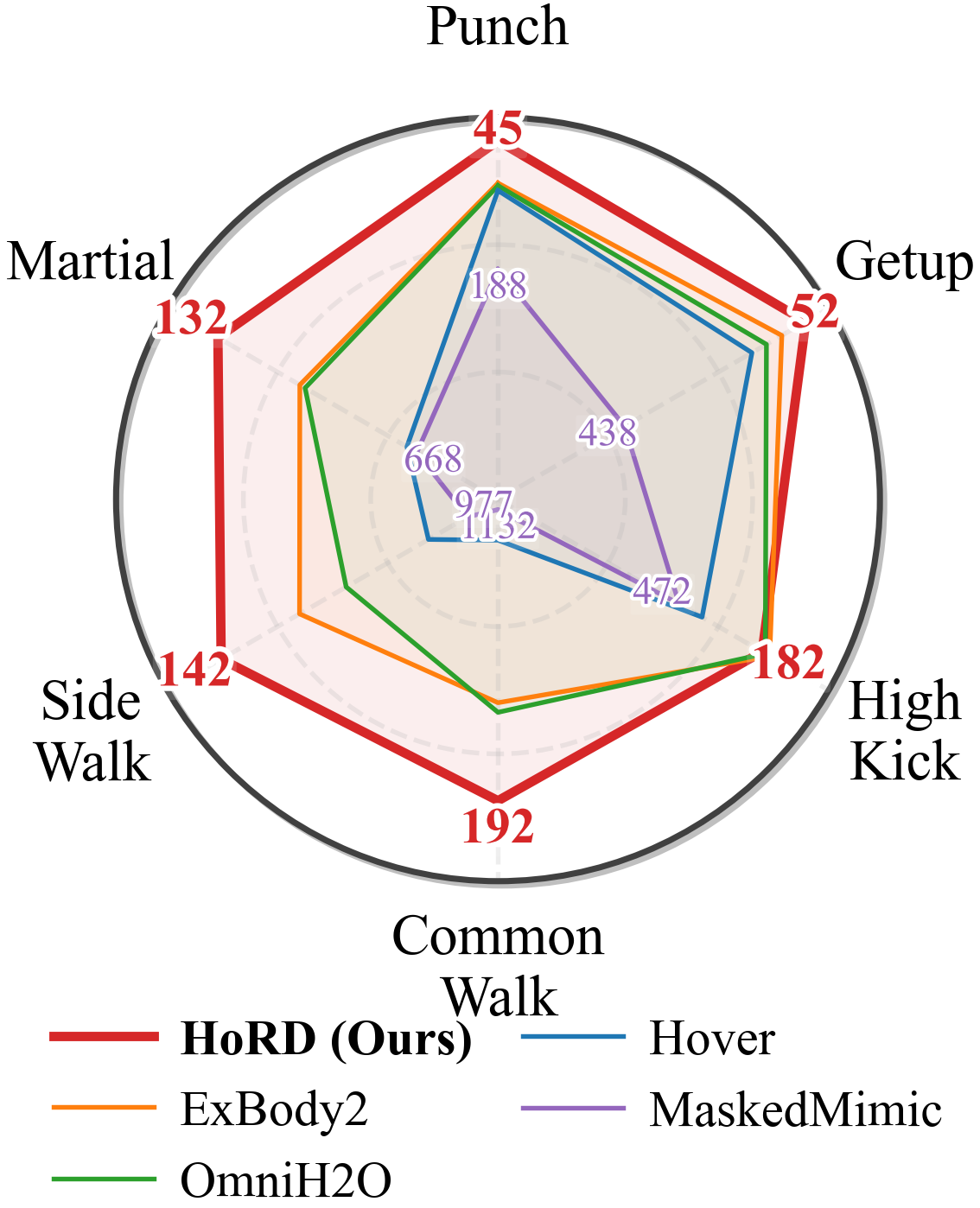}
        \caption{$\mathbf{E_{\text{g-mpjpe}}}\downarrow$(mm)}
        \label{fig:radar-eg}
    \end{subfigure}
    \hfill
    \begin{subfigure}[t]{0.49\linewidth}
        \centering
        \includegraphics[width=\linewidth]{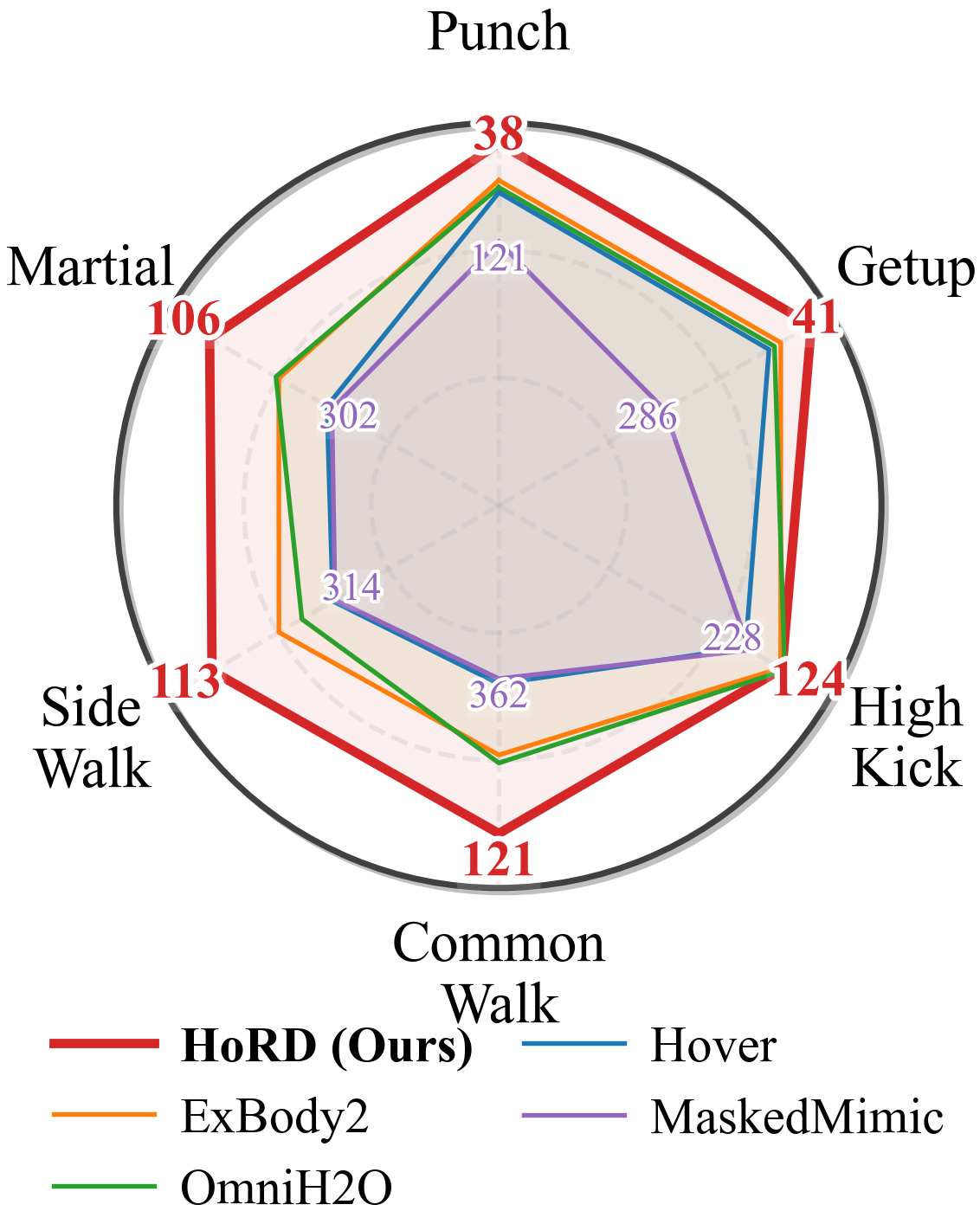}
        \caption{$\mathbf{E_{\text{mpjpe}}}\downarrow$(mm)}
        \label{fig:radar-mpjpe}
    \end{subfigure}
    \vspace{-0.5em}
    \caption{Pose estimation errors comparing HoRD and ExBody2 across six representative motions.}
    \label{fig:radar}
\end{figure}
\vspace{-0.5em}

Figure~\ref{fig:motion} visualizes representative rollouts of HoRD on these motions in the unseen Genesis environment. Across all cases, HoRD produces temporally stable, human-like trajectories while maintaining balance and task completion. In particular, challenging motions such as high kick, which involve rapid center-of-mass shifts and intermittent ground contact, remain stable without collapse. 

Figure~\ref{fig:radar} reports pose errors comparing HoRD with the baselines. HoRD achieves lower errors on most motions in both $\mathbf{E_{\text{g-mpjpe}}}$ and $\mathbf{E_{\text{mpjpe}}}$, with the clearest improvements on common walk, side walk, and martial. These cases emphasize sustained tracking under long-horizon coordination and asymmetric contacts, where small deviations can accumulate over time. Overall, the per-motion breakdown suggests that HoRD’s advantage is most pronounced on motions that require consistent whole-body coordination over longer horizons.

\subsection{Ablation Studies}
\label{subsec:ablations}

\begin{table*}[t]
\centering

\caption{Ablation study of HoRD across different environments.
We analyze the contribution of domain randomization (D) and the HCDR module (H). Note that without SSJR (w/o S), there will be no conversion from motion to torque, and the \textbf{Succ.} will be $<$ 10 for all cases.}
\vspace{-0.5em}
\small
\setlength{\tabcolsep}{5.1pt}
\renewcommand{\arraystretch}{1.1}
\begin{tabular}{lccc|ccc|ccc|ccc}
\toprule
& \multicolumn{3}{c|}{\textcolor{trainhdr}{\textbf{IsaacLab (ID)}}}
& \multicolumn{3}{c|}{\textcolor{trainhdr}{\textbf{IsaacLab + DR (ID)}}}
& \multicolumn{3}{c|}{\textcolor{testhdr}{\textbf{Genesis (OOD)}}}
& \multicolumn{3}{c}{\textcolor{testhdr}{\textbf{Genesis + DR (OOD)}}} \\
\cmidrule(lr){2-4} \cmidrule(lr){5-7} \cmidrule(lr){8-10} \cmidrule(lr){11-13}

\textbf{Method}
& \textbf{Succ.} & $\mathbf{E_{\text{g-mpjpe}}}$ & $\mathbf{E_{\text{mpjpe}}}$
& \textbf{Succ.} & $\mathbf{E_{\text{g-mpjpe}}}$ & $\mathbf{E_{\text{mpjpe}}}$
& \textbf{Succ.} & $\mathbf{E_{\text{g-mpjpe}}}$ & $\mathbf{E_{\text{mpjpe}}}$
& \textbf{Succ.} & $\mathbf{E_{\text{g-mpjpe}}}$ & $\mathbf{E_{\text{mpjpe}}}$ \\
& (\%)$\uparrow$ & (mm)$\downarrow$ & (mm)$\downarrow$
& (\%)$\uparrow$ & (mm)$\downarrow$ & (mm)$\downarrow$
& (\%)$\uparrow$ & (mm)$\downarrow$ & (mm)$\downarrow$
& (\%)$\uparrow$ & (mm)$\downarrow$ & (mm)$\downarrow$ \\
\midrule

HoRD
& 90.7 & 102 & 76
& 88.4 & 124 & 87
& 86.0 & 162 & 96
& 84.4 & 171 & 108 \\

HoRD w/o D
& 79.8 & 186 & 136
& $<$10 & 832 & 286
& $<$10 & 838 & 316
& $<$10 & 838 & 287 \\

HoRD w/o H
& 91.3 & 137 & 97
& 70.5 & 341 & 182
& $<$10 & 846 & 324
& $<$10 & 433 & 234 \\

\bottomrule
\end{tabular}

\label{tab:ablation}
\vspace{-0.5em}
\end{table*}

Ablation studies (Table~\ref{tab:ablation}) reveal complementary roles for domain randomization and HCDR. Removing domain randomization (HoRD w/o D) causes severe performance collapse under test-time perturbations or simulator changes, confirming its necessity for robust generalization beyond the training distribution. 

Removing HCDR (HoRD w/o H) maintains competitive performance on the training distribution but degrades sharply under distribution shift, with near-zero success in Genesis, highlighting its role as a learned dynamics adaptation mechanism that enables online adaptation to latent changes in simulator parameters and contact behavior. Further implementation details, diagnostic metrics, and training convergence analyses are provided in Appendix~\ref{subsec:training_distillation}, Appendix~\ref{subsec:test_metrics}, and Appendix~\ref{subsec:training_curves}.


\subsection{Online Adaptation on Different Terrain Surface}

\begin{table}[t]
\centering
\caption{
Online adaptation performance on different terrain surfaces. 
}
\vspace{-0.5em}
\small
\setlength{\tabcolsep}{3.7pt}
\renewcommand{\arraystretch}{1}
\begin{tabular}{lccc}
\toprule
\textbf{Terrain} 
\textbf{Method} & \textbf{Succ.} (\%)$\uparrow$ & $\mathbf{E_{\text{g-mpjpe}}}$ (mm)$\downarrow$ & $\mathbf{E_{\text{mpjpe}}}$(mm)$\downarrow$ 
\\
\midrule
Flat ground        & 86.2 & 167 & 102 \\
Smooth slope   & 85.8 & 171 & 106 \\
Rough slope & 84.2 & 180 & 112 \\
\bottomrule
\end{tabular}

\label{tab:terrain}
\vspace{-2.5em}
\end{table}

We evaluate terrain robustness in the Genesis simulator under zero-shot transfer. We consider three terrain settings of increasing difficulty: flat ground, smooth slope, and rough slope. We visualize terrain rollouts in Appendix~\ref{subsec:terrain_vis}.

As shown in Table~\ref{tab:terrain}, HoRD maintains strong performance across all terrain conditions. On flat terrain, HoRD achieves an $86.2\%$ success rate with low tracking error. As terrain difficulty increases, performance degrades gracefully, specifically, $85.8\%$ on smooth slopes and $84.2\%$ on rough slopes, closely matching its performance on flat terrian setting.

Qualitative results (Appendix Fig.~\ref{fig:terrain}) further illustrate this trend. HoRD produces stable and continuous motions across all terrains, adapting foot placement and body posture to compensate for incline, whereas baselines frequently lose balance or exhibit oscillatory behavior. These results demonstrate that HoRD generalizes beyond flat-ground assumptions and provides robustness control under terrain-induced dynamics variations.

\subsection{Disturbance Recovery}

To further evaluate robustness under unexpected physical disturbances, we test HoRD in the same zero-shot Genesis environment above, while injecting random external pushes during execution. These disturbances are applied intermittently at random time intervals, perturbing the humanoid’s root velocity in the world frame and simulating sudden external forces such as collisions or human interference.

\begin{figure}[t]
    \centering
    \includegraphics[width=\linewidth]{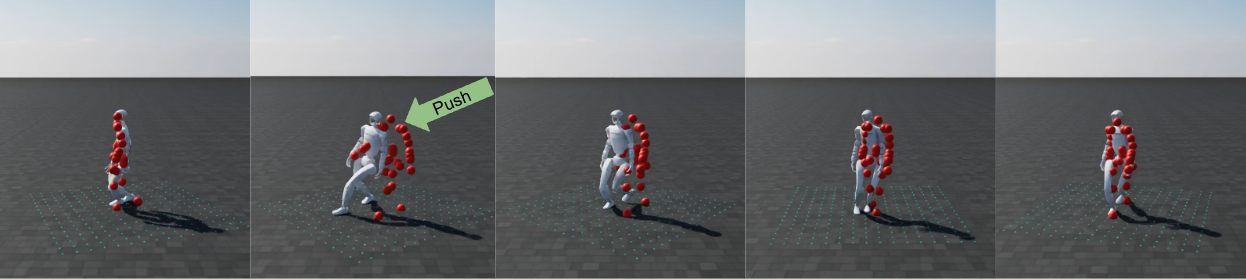}
    \caption{
Online adaptation experiment shows that a lateral push is applied mid-execution (second frame), perturbing the humanoid’s motion. HoRD rapidly re-stabilizes and resumes the intended trajectory (red markers).
}
    \label{fig:push}
\vspace{-0.5em}
\end{figure}

Figure~\ref{fig:push} shows a representative recovery sequence, where a push is applied during walking motion execution. Despite the abrupt disturbance, HoRD quickly absorbs the impact, re-stabilizes the body, and resumes the intended motion without falling or task failure. This behavior is consistently observed across evaluation rollouts. HoRD recovers successfully in 85.2\% of disturbed executions, outperforming strong baselines such as OmniH2O (71.8\%) and ExBody2 (72.1\%). These results show the robustness of HoRD under disturbance and indicate that HoRD performs closed-loop execution rather than open-loop motion replay.

\section{Discussion and Future Works}
\label{sec:discussion}

\vspace{-0.5em}
\paragraph{Data and behavior coverage.}
Our experiments are based on AMASS-derived motions. Because the humanoid differs from humans in mass distribution and torque limits, some clips are effectively infeasible and, despite basic filtering, can introduce gradient interference with more plausible motions. Object-centric behaviors that rely on environmental props (e.g., chairs, external supports) are also largely outside our current scope. More careful feasibility-aware curation and extending HoRD to richer scenes with explicit object interactions are natural next steps.

\vspace{-0.5em}
\paragraph{Learning efficiency and optimization cost.}
The expert policy is learned via on-policy RL in a broad domain-randomized setting, which remains computationally expensive and sensitive to reward design, observation noise, and randomization ranges. Learning policies that cover a wide dynamics distribution on AMASS-scale data requires days of training on a single high-end GPU, and overly aggressive randomization can create competing gradients across motions. Future work will explore more sample-efficient learning schemes and automated calibration of domain randomization to better balance performance on the training distribution and cross-domain generalization.

\vspace{-0.5em}
\paragraph{Long-horizon recovery and goal-aware control.}
HoRD demonstrates strong robustness to perturbations and emergent fall-recovery behaviors, but its learned policy remains conditioned primarily on short-horizon sparse inputs. A natural extension is to explore lightweight goal or waypoint-conditioned variants. Such extensions can be incorporated into SSJR and learned within the HoRD framework, providing a principled path toward richer recovery and higher-level whole-body behaviors.

\section{Conclusion}
\label{sec:conclusion}
We introduced HoRD, a framework mapping sparse motion commands to robust torque-level humanoid control. By combining standardized sparse-joint representation with history-conditioned dynamics adaptation, HoRD learns stable whole-body behaviors that generalize across substantial dynamics variation. HoRD significantly outperforms baselines in tracking fidelity, stability, and zero-shot transfer via data-efficient distillation from a domain-randomized expert. HoRD shows that lightweight learned policies with standardized representations can bridge high-level planners and deployment-ready control.


\section*{Impact Statement}

This work improves robustness in torque-level humanoid control by introducing a learning framework that (i) decouples high-level motion intent from low-level actuation via standardized representations, (ii) operates from sparse motion commands that can be produced from diverse sources such as vision--language-action models, video, or wearables, and (iii) generalizes under distribution shifts in physical dynamics or simulators through online adaptation from recent state--action history. Such capability may benefit embodied AI research and, in the longer term, assistive and service robotics by reducing the engineering and compute overhead of re-training or re-tuning controllers across robots and deployment conditions, while the released large-scale trajectory dataset and evaluation scripts support reproducible benchmarking.

However, more capable and transferable humanoid control policies raise safety considerations if deployed without sufficient verification. In particular, real-world dynamics changes that are not directly observed (e.g., mass distribution, friction, actuator delay) can cause small model mismatches to compound into large tracking errors and destabilize torque control over long horizons. We therefore emphasize responsible deployment: systems built on this technique should undergo extensive validation beyond simulation, incorporate safety constraints and fail-safes (e.g., action limiting, emergency stops, and monitoring), and maintain clear usage boundaries with human oversight on physical platforms. Robustness may also vary across embodiments, sensing configurations, and operating conditions; future work should audit reliability across diverse settings and strengthen evaluation protocols for sim-to-real transfer and safety.

\bibliographystyle{icml2026}
\bibliography{refs}

\newpage
\appendix
\onecolumn
\clearpage
\appendix

\section{Additional Results}
\label{app:additional_results}

\subsection{Qualitative Comparisons}
\label{subsec:qualitative_comparisons}

\begin{figure}[H]
    \centering
    \begin{subfigure}[b]{0.49\textwidth}
        \centering
        \includegraphics[width=\textwidth]{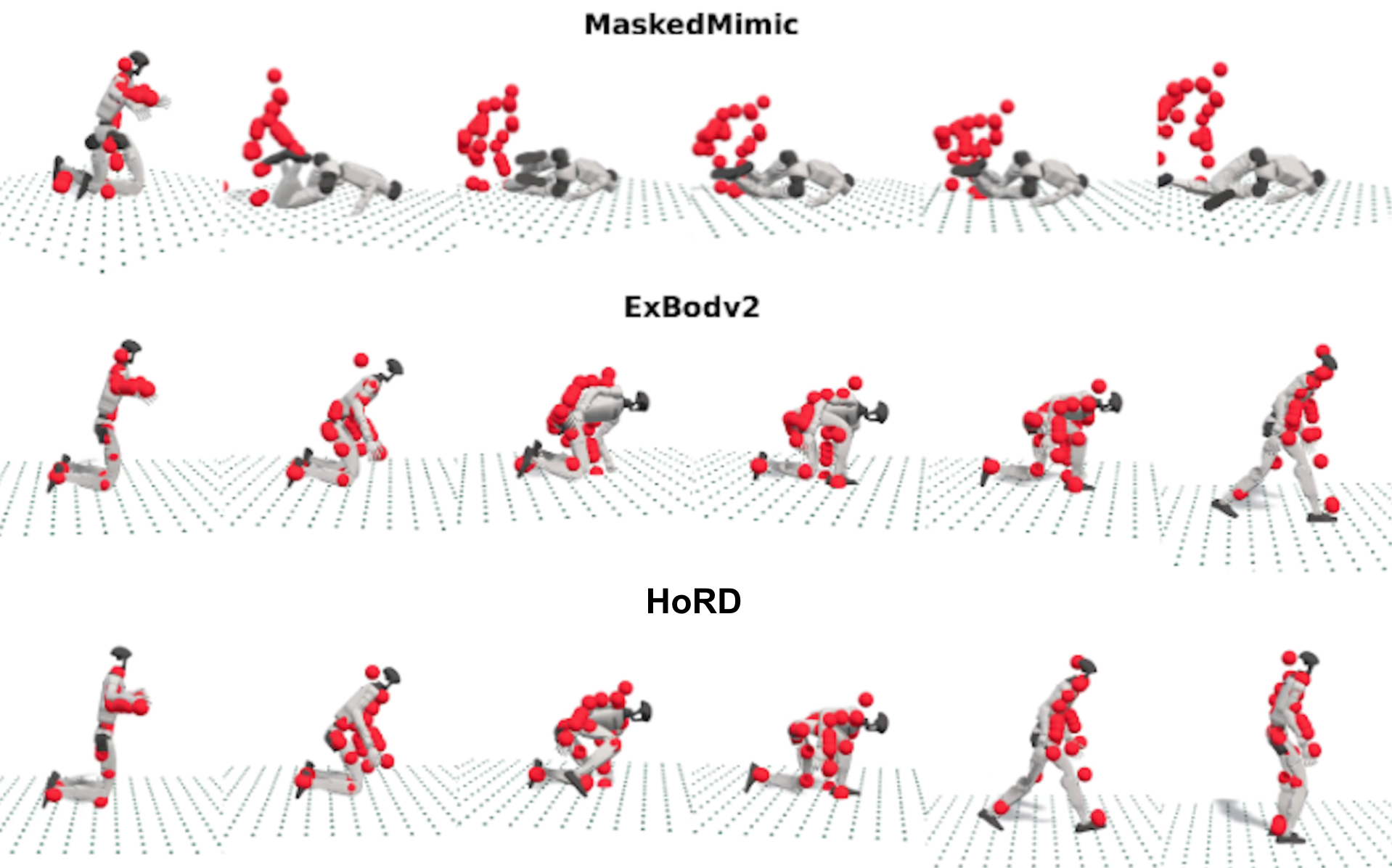}
        \caption{Punch}
    \end{subfigure}
    \hfill
    \begin{subfigure}[b]{0.49\textwidth}
        \centering
        \includegraphics[width=\textwidth]{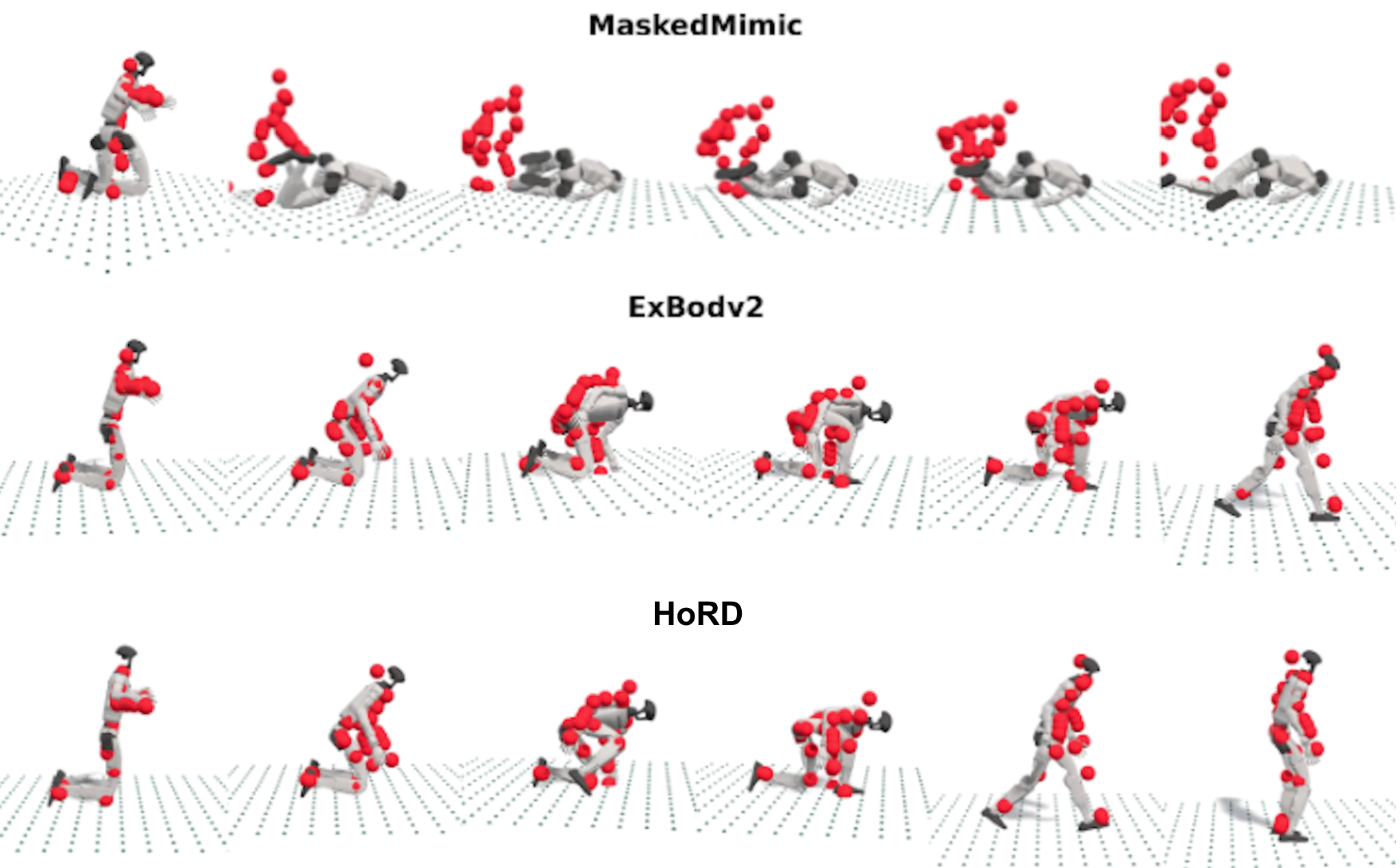}
        \caption{Get Up}
    \end{subfigure}

    \vspace{0.8em}

    \begin{subfigure}[b]{0.49\textwidth}
        \centering
        \includegraphics[width=\textwidth]{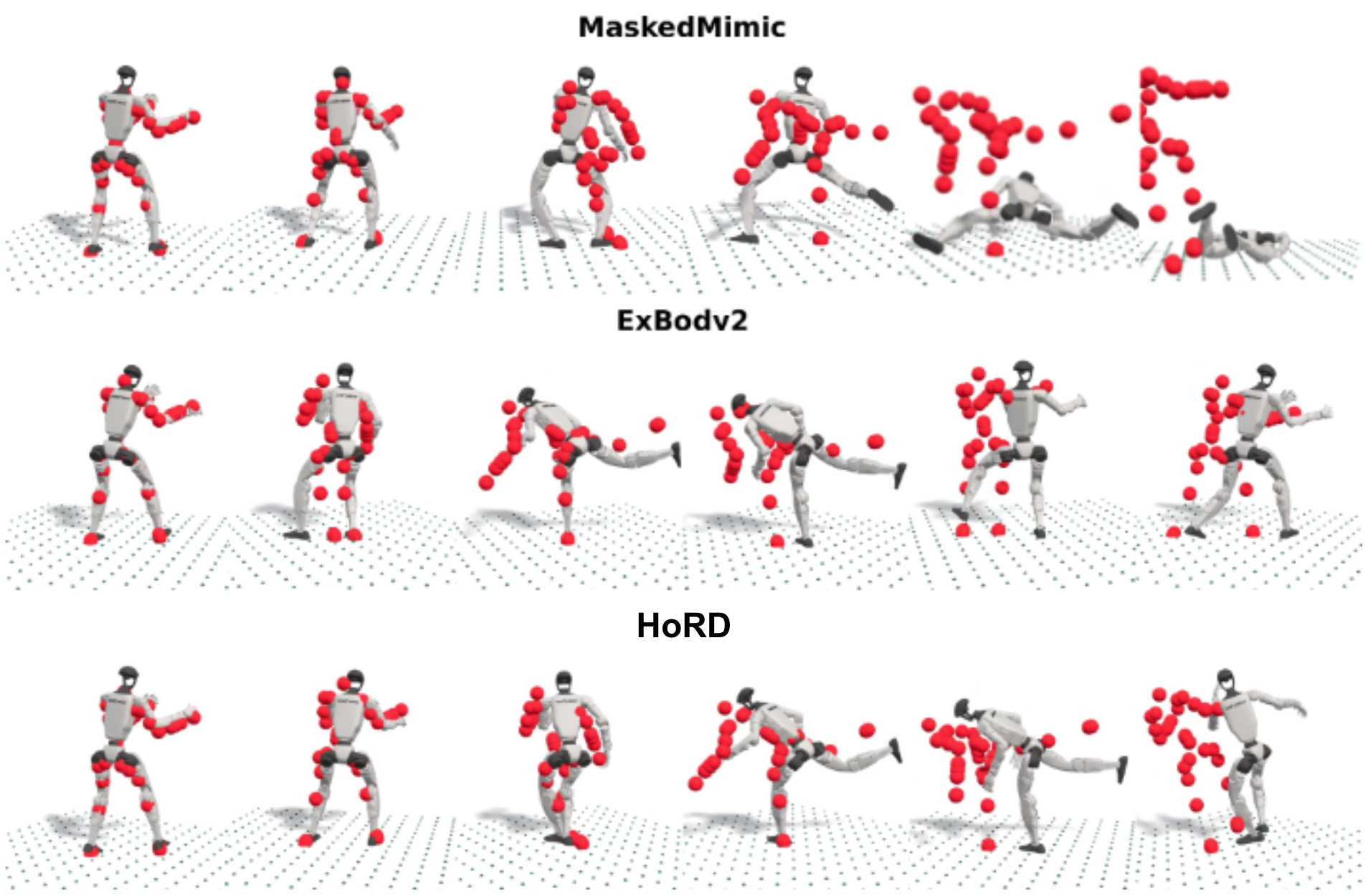}
        \caption{High Kick}
    \end{subfigure}
    \hfill
    \begin{subfigure}[b]{0.49\textwidth}
        \centering
        \includegraphics[width=\textwidth]{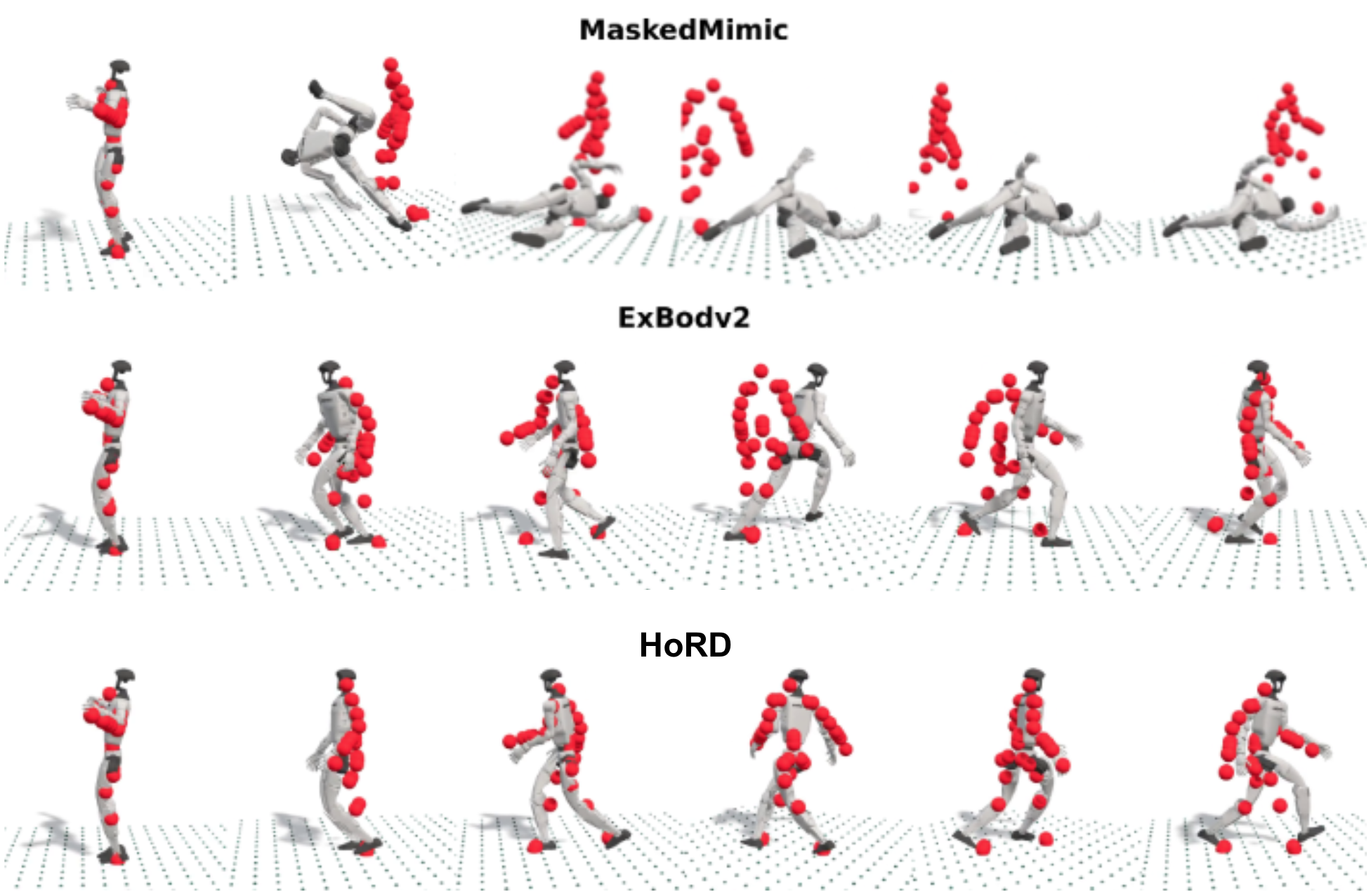}
        \caption{Common Walk}
    \end{subfigure}

    \vspace{0.8em}

    \begin{subfigure}[b]{0.49\textwidth}
        \centering
        \includegraphics[width=\textwidth]{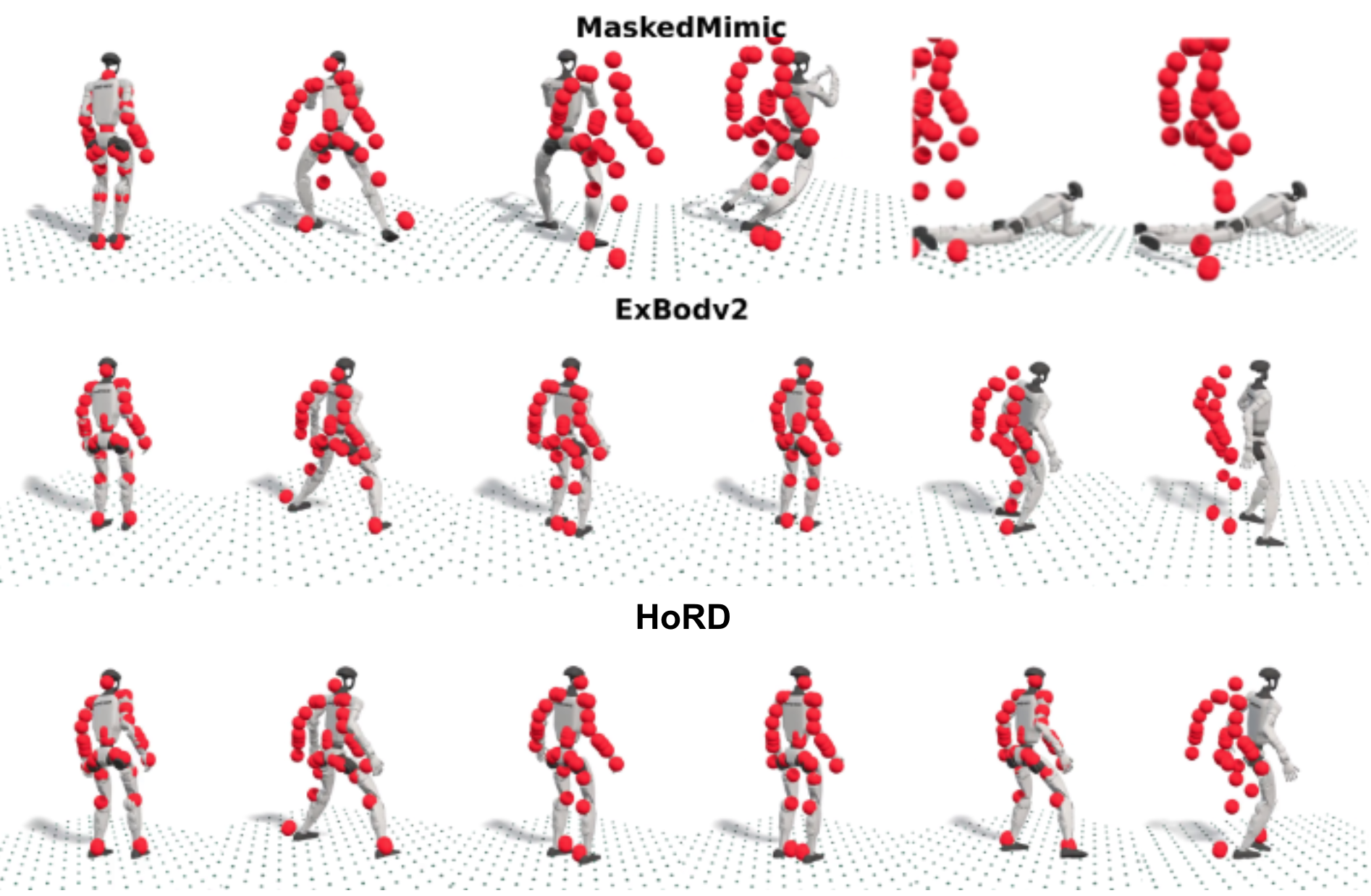}
        \caption{Side Walk}
    \end{subfigure}
    \hfill
    \begin{subfigure}[b]{0.49\textwidth}
        \centering
        \includegraphics[width=\textwidth]{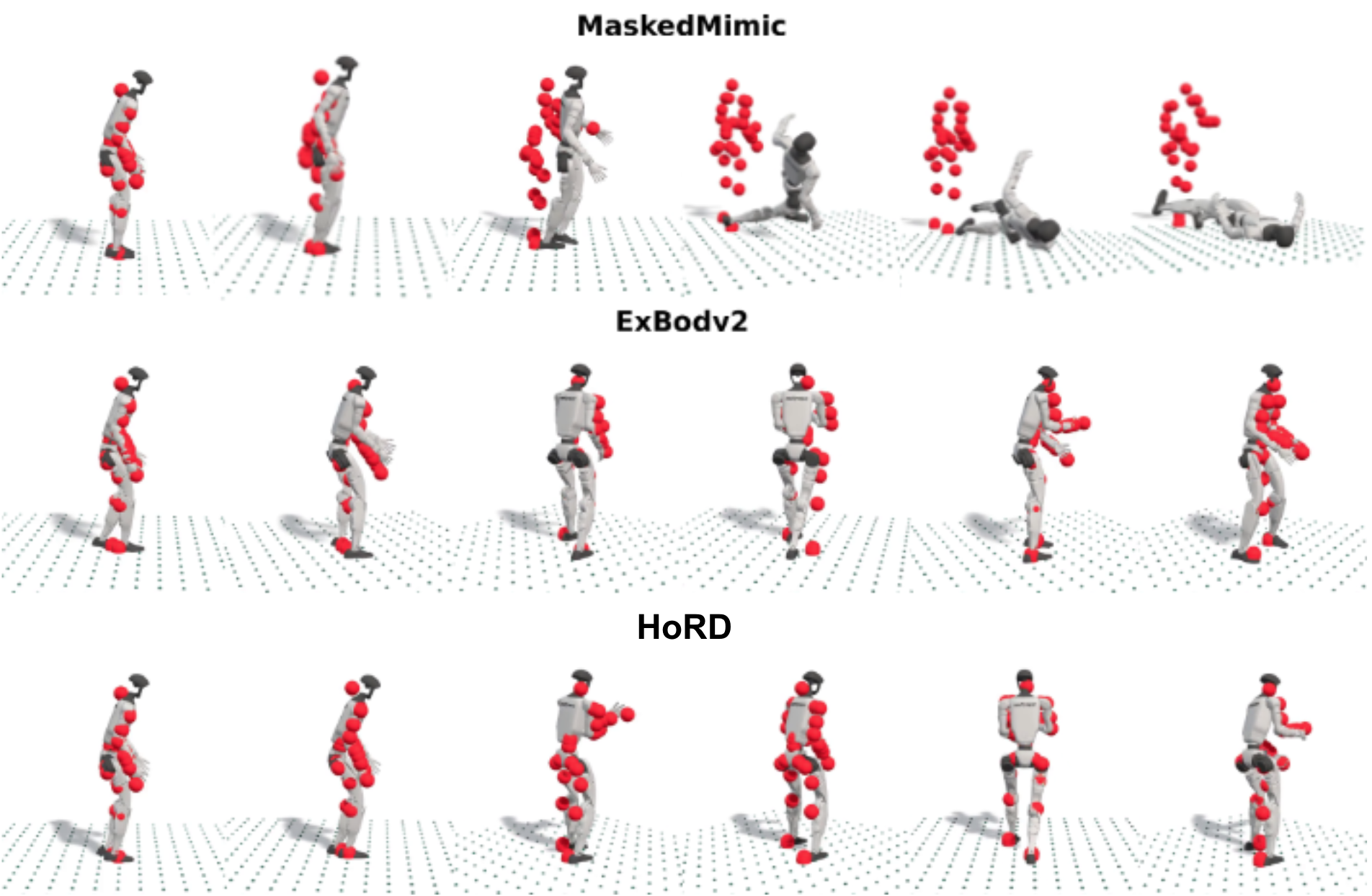}
        \caption{Martial Arts}
    \end{subfigure}

    \caption{Qualitative comparisons across 6 different motions.}
    \label{fig:qualitative_comparisons}
\end{figure}

Fig.~\ref{fig:qualitative_comparisons} provides qualitative visualizations corresponding to the quantitative results (Figure~\ref{fig:radar}) in the Section~\ref{subsec:motions}. Specifically, we visualize representative execution trajectories of HoRD together with the strongest competing baseline and the weakest-performing baseline identified, MaskedMimic and ExBody2.

The examples are intended to illustrate how differences in tracking errors reported by the quantitative metrics manifest in the resulting motion executions, and to provide additional visual context for the comparative results. In several cases, MaskedMimic exhibits early instability or tracking failure, while ExBody2 shows noticeable drift or delayed recovery. In contrast, HoRD maintains closer alignment with the reference motion throughout the execution.

\subsection{Test Set Evaluation Metrics}
\label{subsec:test_metrics}

We present comprehensive evaluation metrics on the test set. These results are obtained from learned policies evaluated in IsaacLab with domain randomization (DR) settings on held-out test motions. 
Table~\ref{tab:ablation_metrics} summarizes the performance comparison across different evaluation metrics for HoRD and its ablation variants.

\begin{table}[h]
    \centering
    \small
    \caption{Comprehensive evaluation metrics comparison across different methods on the test set. 
    All policies are evaluated in IsaacLab with domain randomization.}
    \label{tab:ablation_metrics}
    \begin{tabular}{lccc}
        \toprule
        \textbf{Metric} & \textbf{HoRD} & \textbf{HoRD w/o D} & \textbf{HoRD w/o M} \\
        \midrule
        Cartesian Error $\downarrow$                & 0.087 & 0.286 & 0.182 \\
        Global Rotation Error $\downarrow$          & 0.369 & 1.594 & 1.011 \\
        Global Translation Error $\downarrow$       & 0.124 & 0.832 & 0.341 \\
        \midrule
        DOF Velocity Reward $\uparrow$              & 0.972 & 0.980 & 0.960 \\
        Key Body Reward $\uparrow$                  & 0.907 & 0.412 & 0.612 \\
        Local Rotation Reward $\uparrow$            & 0.833 & 0.274 & 0.507 \\
        Root Angular Velocity Reward $\uparrow$     & 0.480 & 0.722 & 0.412 \\
        Root Velocity Reward $\uparrow$             & 0.954 & 0.878 & 0.904 \\
        \bottomrule
    \end{tabular}
\end{table}

\subsection{Discussion on Extra Baselines}
\label{subsec:method_comparison}

We compare HoRD with existing methods across three deployment-relevant capabilities, as summarized in Table~\ref{tab:method_comparison}: unified skill coverage, sparse-command generalization, and explicit online dynamics adaptation. HoRD supports unified policy learning over heterogeneous skills while remaining robust when only sparse or partial keypoint-level commands are available. In addition, HoRD incorporates an explicit online dynamics adaptation mechanism through HCDR's history-conditioned representation, enabling the policy to infer latent dynamics variations and adjust its strategy accordingly at test time. In contrast, prior approaches typically emphasize only a subset of these capabilities, for example focusing on unified motion priors or partial-constraint training without dedicated dynamics adaptation.



\begin{table*}[t]
\centering
\caption{Comparison of representative humanoid control frameworks in terms of three deployment-relevant capabilities: (i) \textbf{Unified Skill Coverage}, indicating whether a single policy can cover diverse heterogeneous skills; (ii) \textbf{Sparse-Command Generalization}, reflecting robustness when only sparse or partial keypoint-level commands are available; and (iii) \textbf{Explicit Online Dynamics Adaptation}, denoting the presence of a dedicated mechanism for adapting to dynamics variations at test time.}
\label{tab:method_comparison}

\begin{tabular}{lccc}
\toprule
Method &
Unified Skill Coverage &
Sparse-Command Generalization &
Explicit Online Dynamics Adaptation \\
\midrule
BumbleBee          & $\times$    & $\times$    & $\times$    \\
HOVER              & $\times$    & \checkmark & $\times$    \\
MaskedMimic        & \checkmark & \checkmark   & $\times$    \\
ASAP               & $\times$    & $\times$    & $\times$    \\
PHC                & $\times$    & $\times$    & $\times$    \\
PULSE              & $\times$    & $\times$    & $\times$    \\
ExBody2            & $\times$    & $\times$    & $\times$    \\
OmniH2O            & \checkmark & $\times$    & $\times$    \\
HoRD (ours)      & \checkmark & \checkmark & \checkmark \\
\bottomrule
\end{tabular}

\vspace{2pt}
{\footnotesize \textbf{Notation:} \checkmark~supported; $\times$~not supported.}
\end{table*}

\subsection{Performance across different terrains}
\label{subsec:terrain_vis}
\begin{wrapfigure}{r}{0.4\linewidth}
    \centering
    \vspace{-3em}
    \includegraphics[width=\linewidth]{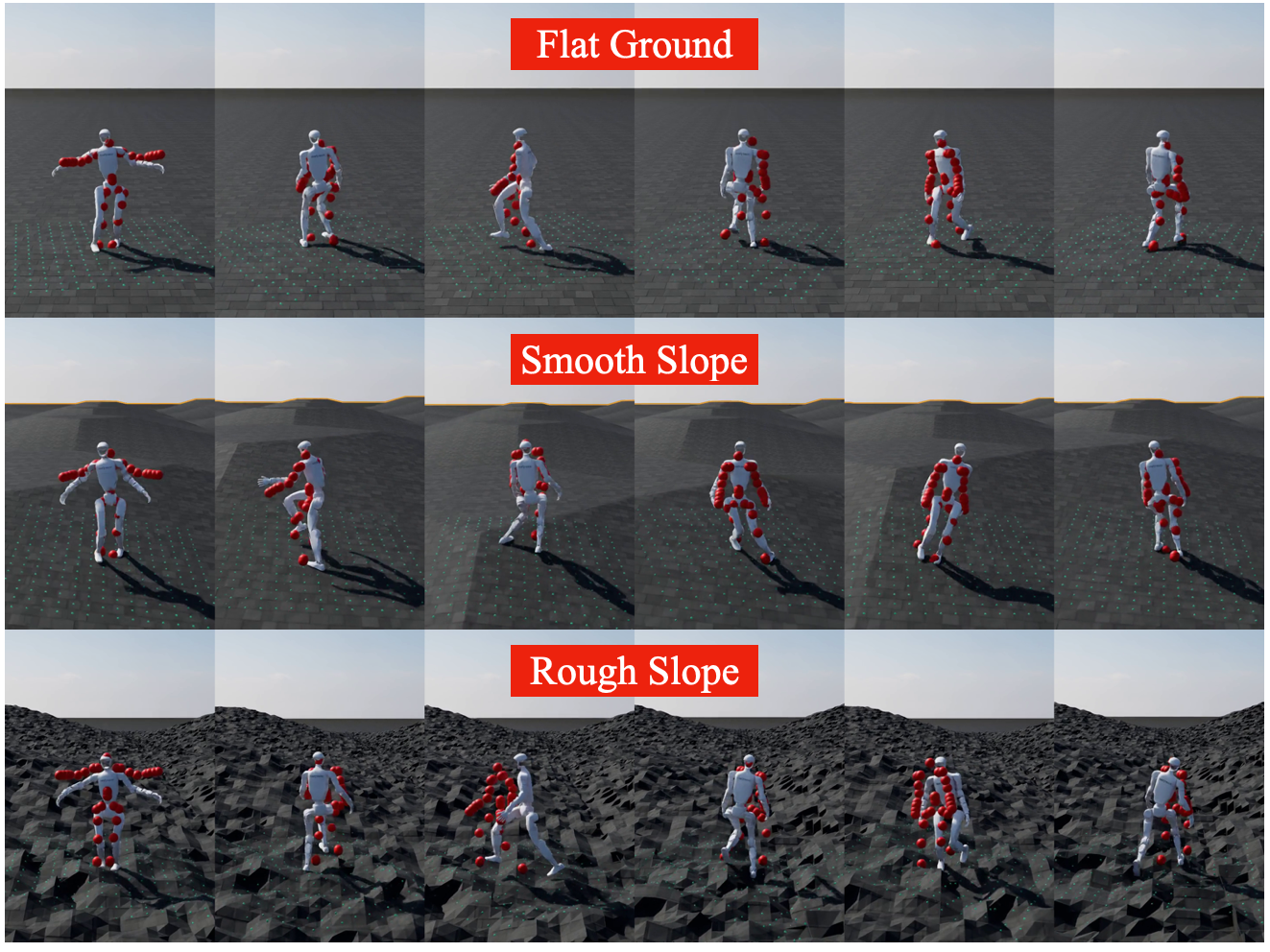}
    \caption{Visualization of HoRD performance across different terrains.}
    \label{fig:terrain}
    \vspace{-3em}
\end{wrapfigure}
Figure~\ref{fig:terrain} visualizes HoRD locomotion across three terrain types: flat ground, a smooth incline, and a rough incline. On flat ground, the robot maintains stable forward motion with a consistent posture. On the smooth slope, HoRD adapts its body pitch and foot placement to climb steadily without noticeable slipping. On the rough slope, it remains robust to uneven contacts, adjusting step timing and stance to preserve balance and traction while ascending.

\subsection{Baseline Training Details}
\label{subsec:baseline_details}

To evaluate the effectiveness of our method, we compare it with four state-of-the-art (SOTA) methods: Maskedmimic, OmniH2O, ExBody2, and HOVER. For fair comparison, we either use the officially released code or closely follow the official implementation when code is unavailable. We adapt each method to the Unitree G1 robot (instead of H1/H1-2). In addition, we keep the training dataset consistent with that used for HoRD. For HOVER, we use unmasked observations during the evaluation to ensure optimal performance.

This section summarizes the training configurations of the baseline methods used for comparison. Whenever possible, we follow the training settings reported in the original papers or official implementations. For methods whose training pipelines are not fully specified or publicly available, we adopt reasonable approximations consistent with prior work and document the key differences explicitly.

All baselines are evaluated under the same test environments and metrics as our method. 

\subsection{SSJR Example}
\label{subsec:SSJR_example}
For reproducibility, we provide a minimal SSJR packet example. It specifies the sequence frame rate (30~fps), duration (10.0~s), number of frames (300), and number of joints (33). The packet stores time-indexed motion fields with fixed tensor dimensions, including global translation $[300,33,3]$, global rotation matrices $[300,33,3,3]$, global rotations in quaternion form $[300,33,4]$, global linear and angular velocities $[300,33,3]$, local joint rotations $[300,29,4]$, root linear and angular velocities $[300,3]$, and joint DoF positions and velocities $[300,29]$. Finally, it optionally includes text tokens (e.g., a brief natural-language description of the motion) to support downstream conditioning. 




\section{Environment Details}
\label{app:environment}

\subsection{RL Settings}
\label{subsec:rl_env}

We provide a detailed training and test environment setting in this subsection.

\paragraph{Observation.}
For privileged observation (teacher policy), we use proprioception including linear velocity, angular velocity, joint position, joint velocity, and last action, as well as task-relevant observation including target joint positions, target keypoint positions, target root translations, and target root rotations in global coordinates. 

Temporal context is provided by stacking observations over multiple timesteps.
The teacher policy consumes observations from the past 5 timesteps, while the student policy uses a longer horizon of 10 timesteps (8-step history with 2-step lookahead).
The resulting proprioceptive observation dimension is 493, and target poses include an 8-step future with 18 dimensions per joint.

\paragraph{Action and actuation model.}
\label{subsec:pd_target}
We operate a torque-actuated simulator of the Unitree G1 humanoid with 29 actuated degrees of freedom (DoFs).
The policy outputs desired joint configurations $\boldsymbol{\alpha}_t$, which are converted into joint torques through a proportional–derivative (PD) controller and applied in simulation.
Concretely, the applied torque at time $t$ is given by
\[
\boldsymbol{\tau}_t = \mathbf{K}_p \bigl(\boldsymbol{\alpha}_t - \mathbf{q}_t\bigr) - \mathbf{K}_d \,\dot{\mathbf{q}}_t ,
\]
where $\boldsymbol{\alpha}_t$ denotes the target joint positions predicted by the policy, and $\mathbf{q}_t$, $\dot{\mathbf{q}}_t$ are the current joint positions and velocities.
Stiffness and damping gains are specified per joint group, with stiffness values of 100 for hips, 200 for knees, 400 for the waist, and 60--90 for shoulders, and damping coefficients ranging from 0.1 to 5.0.

\paragraph{Termination.}
Each episode follows the duration of a target motion clip. An episode terminates early if the global tracking error exceeds a threshold of $0.5$\,m, indicating irreversible divergence.

\subsection{Deployment}
\label{subsec:deployment}

The policy runs at 50\,Hz, while the simulator operates at 200\,Hz with a decimation factor of 4, corresponding to four simulation steps per control action. Policy inference and actuation are integrated through standard simulation interfaces in IsaacLab.

\subsection{Metrics}
\label{subsec:metrics}

We evaluate learned policies using task-level success, trajectory tracking errors, and several reward-aligned kinematic terms. Unless otherwise stated, expectations are taken over time and over evaluation episodes.

\paragraph{Strict tracking success.}
Our primary metric is the strict success rate. For an episode $i$, let $\mathrm{GTE}_t^{(i)}$ denote the global translation error at time $t$ (defined below). The episode is counted as successful if the per-frame error never exceeds the threshold $\varepsilon_{\mathrm{GTE}}{=}0.5$\,m,
\[
\text{success}^{(i)} = \mathbb{I}\Big[\max_{t \le T^{(i)}} \mathrm{GTE}_t^{(i)} < \varepsilon_{\mathrm{GTE}}\Big],
\]
and the strict success rate over $N$ episodes is
\[
\text{success rate} = \frac{1}{N}\sum_{i=1}^{N} \text{success}^{(i)}.
\]

\paragraph{Cartesian position error (cartesian\_err, MPJPE).}
To measure local trajectory fidelity, we compute the mean per-joint position error (MPJPE) in the local frame, after subtracting the initial root translation:
\[
\text{cartesian\_err}
= \mathbb{E}\Big[\big\| ( \hat{p}^{\text{ref}}_t - \hat{p}^{\text{ref}}_0 ) - ( \hat{p}_t - \hat{p}_0 ) \big\|_2 \Big],
\]
where $\hat{p}_t$ and $\hat{p}^{\text{ref}}_t$ are local-frame joint positions for the executed and reference motions, respectively.

\paragraph{Global translation error (gt\_err, Global MPJPE).}
Global alignment is quantified by the RMS error between reference and executed joint positions in world coordinates:
\[
\text{gt\_err}
= \mathbb{E}\Big[\big\| p^{\text{ref}}_t - p_t \big\|_2\Big],
\]
and we write $\mathrm{GTE}_t = \| p^{\text{ref}}_t - p_t \|_2$ for the instantaneous error used in the strict success criterion.

\paragraph{Global rotation error (gr\_err).}
We also report the RMS angular error between reference and executed global orientations:
\[
\text{gr\_err} = \mathbb{E}\!\big[\Delta q(gr_t, gr^{\text{ref}}_t)\big], \qquad
\text{gr\_err\_degrees} = \text{gr\_err} \cdot \frac{180}{\pi},
\]
where $\Delta q(\cdot,\cdot)$ denotes the angular distance between two unit quaternions.

\paragraph{Reward-aligned kinematic terms.}
For completeness, we additionally report several exponentiated tracking rewards that HCDR components of the teacher's RL objective, which are not used as primary evaluation metrics, including DoF velocity tracking, key-body Cartesian tracking, local joint rotation consistency, root linear velocity tracking, and root angular velocity tracking:

\begin{align*}
\text{dv\_rew} &= \exp\!\left(-c_{\text{dv}} \,\big\| \dot{q}_t - \dot{q}^{\text{ref}}_t \big\|_2^2 \right) ,\\
\text{kb\_rew} &= \exp\!\left(-c_{\text{kb}} \,\big\| b_t - b^{\text{ref}}_t \big\|_2^2 \right),\\
\text{lr\_rew} &= \exp\!\left(-c_{\text{lr}} \,\Delta q(lr_t, lr^{\text{ref}}_t)\right),\\
\text{rv\_rew} &= \exp\!\left(-c_{\text{rv}} \,\big\| v^{\text{root}}_t - v^{\text{root,ref}}_t \big\|_2^2 \right),\\
\text{rav\_rew} &= \exp\!\left(-c_{\text{rav}} \,\big\| \omega^{\text{root}}_t - \omega^{\text{root,ref}}_t \big\|_2^2 \right).
\end{align*}
Here $\dot{q}_t$ denotes joint velocities, $b_t$ key-body points (e.g., hands, feet, head), $lr_t$ local joint rotations, and $v^{\text{root}}_t$, $\omega^{\text{root}}_t$ the root linear and angular velocities; $c_{\text{dv}}, c_{\text{kb}}, c_{\text{lr}}, c_{\text{rv}}, c_{\text{rav}}$ are positive scaling constants shared across methods.

\paragraph{Legacy aggregate metrics.}
For comparability to prior work, some tables additionally report legacy aggregate metrics used in MaskedMimic, namely trajectory error $E_{\text{traj}}$ and torque smoothness $S_\tau$.

\section{Training Details}
\label{app:training}

\subsection{Reward Design}
\label{subsec:reward_design}

The reward design for our tracking policy includes task rewards, penalties, and regularization terms. The main reward components are:

\begin{itemize}
\item \textbf{Task rewards:} Global translation (GT) weight=0.5, Global rotation (GR) weight=0.4, Global velocity (GV) weight=0.1, Global angular velocity (GAV) weight=0.1, Root height (RH) weight=0.2
\item \textbf{Penalties:} Global rotation constraint (GR$_c$)=-10, Global translation constraint (GT$_c$)=-100, Root tracking constraint (RT$_c$)=-120, Root height constraint (RH$_c$)=-100
\item \textbf{Regularization:} Power penalty = $5\!\times\!10^{-6}$
\end{itemize}

The reward function encourages accurate motion tracking while maintaining physical feasibility and smooth control.

\subsection{Domain Randomization}
\label{subsec:domain_randomization}

To encourage cross-domain generalization and zero-shot transfer, we apply structured domain randomization at the beginning of each episode. The detailed domain randomization setup is summarized in Table~\ref{tab:domain-rand}.

\begin{table}[t]
\centering
\caption{Domain randomization ranges used for both training and evaluation.}
\label{tab:dr_ranges}
\begin{tabular}{lll}
\toprule
Aspect & Parameter & Range / distribution \\
\midrule
Control delay &
$d_{\text{ctrl}}$ &
$\{0,1,2,3\}$ physics steps \\
Body mass scale &
$s_m$ &
$\mathcal{U}[0.9, 1.1]$ \\
Joint parameter scale &
$s_j$ &
$\mathcal{U}[0.9, 1.1]$ \\
Actuator gain scale &
$s_a$ &
$\mathcal{U}[0.9, 1.1]$ \\
Gravity (downwards) &
$g_z$ &
$\mathcal{U}[9.7, 9.9]\ \mathrm{m/s^2}$ \\
External push interval &
$\Delta t_{\text{push}}$ &
$\mathcal{U}[5, 10]\ \mathrm{s}$ \\
\bottomrule
\end{tabular}

\label{tab:domain-rand}
\vspace{-2em}
\end{table}

This randomization envelope is shared across HoRD and all baselines to ensure fair comparison.

\subsection{RL Hyperparameters}
\label{subsec:rl_hyperparams}

The RL training follows standard PPO. We provide the detailed training hyperparameters in Table~\ref{tab:hyperparams}.

\begin{table}[h]
\small
\centering
\caption{Key training hyperparameters.}
\label{tab:hyperparams}

\begin{tabular}{ll}
\toprule
\textbf{Category} & \textbf{Values} \\
\midrule
\textbf{Training setup} 
& Env count: 4,096; Max steps: $1\!\times\!10^{10}$; Seed: 0 \\
& Simulator FPS: 200; Decimation: 4; Mixed terrains (7 levels) \\
\midrule
\textbf{Optimization} 
& PPO $\gamma$=$0.99$, $\tau$=$0.95$, clip $\epsilon$=$0.2$ \\
& Actor LR=$2\!\times\!10^{-5}$, Critic LR=$1\!\times\!10^{-4}$ \\
& Batch size=16,384, Mini epochs=1, Grad clip=50.0 \\
\midrule
\textbf{Model} 
& Transformer: 6 layers, 8 heads, dim=512, FF=1,024 \\
& Actor log std = -2.9; Dropout = 0; Activation = ReLU \\
& Critic: 4-layer MLP (1,024 hidden units each) \\
\midrule
\textbf{Observation encoding} 
& Self obs dim = 493; Hist steps = 8 \\
& Target poses: 8-step future, 18 dims per joint \\
& Terrain encoding: 256-256 MLP with normalization \\
\midrule
\textbf{Reward shaping} 
& Weights: GT=0.5, GR=0.4, GV=0.1, GAV=0.1, RH=0.2 \\
& Penalties: GR$_c$=$-10$, GT$_c$=$-100$, RT$_c$=$-120$, RH$_c$=$-100$ \\
& Power penalty = $5\!\times\!10^{-6}$ \\
\midrule
\textbf{Control and robot} 
& Actions: 29 DoFs, Proportional PD control \\
& Stiffness: hips=100, knees=200, waist=400, shoulders=60--90 \\
& Damping: 0.1--5.0 depending on joint group \\
\midrule
\textbf{Terrain} 
& Terrain types: [smooth slope, rough slope, stairs up, stairs down, discrete, stepping, poles, flat] \\
& Terrain proportions: [0.25, 0.25, 0., 0., 0., 0., 0., 0.5] \\
\bottomrule
\vspace{-2em}
\end{tabular}
\end{table}

\subsection{Motion Dataset Processing}
\label{subsec:dataset_processing}

Following prior work, we first retarget SMPL-format human motion sequences from the AMASS dataset into robot-specific representations, including global translations and joint axis-angle rotations. Given that AMASS contains a wide range of motions, such as crawling and climbing, we perform an additional data cleaning step to ensure quality. We adopt the data filtering procedures described in PHC and MaskedMimic to clean the dataset, and split the dataset following MaskedMimic's data splitting protocol.

Motion sequences are segmented into clips of 2--10 seconds and retargeted to the humanoid morphology. Clips are discarded if retargeting violates joint limits or induces more than $5\%$ persistent contact penetration. All remaining clips are downsampled to 30\,Hz and converted into sparse key-joint trajectories using the proposed SSJR interface.

\subsection{Training and Distillation}
\label{subsec:training_distillation}

\begin{figure}[t]
    \centering
    \begin{subfigure}[b]{\textwidth}
        \centering
        \includegraphics[width=\textwidth]{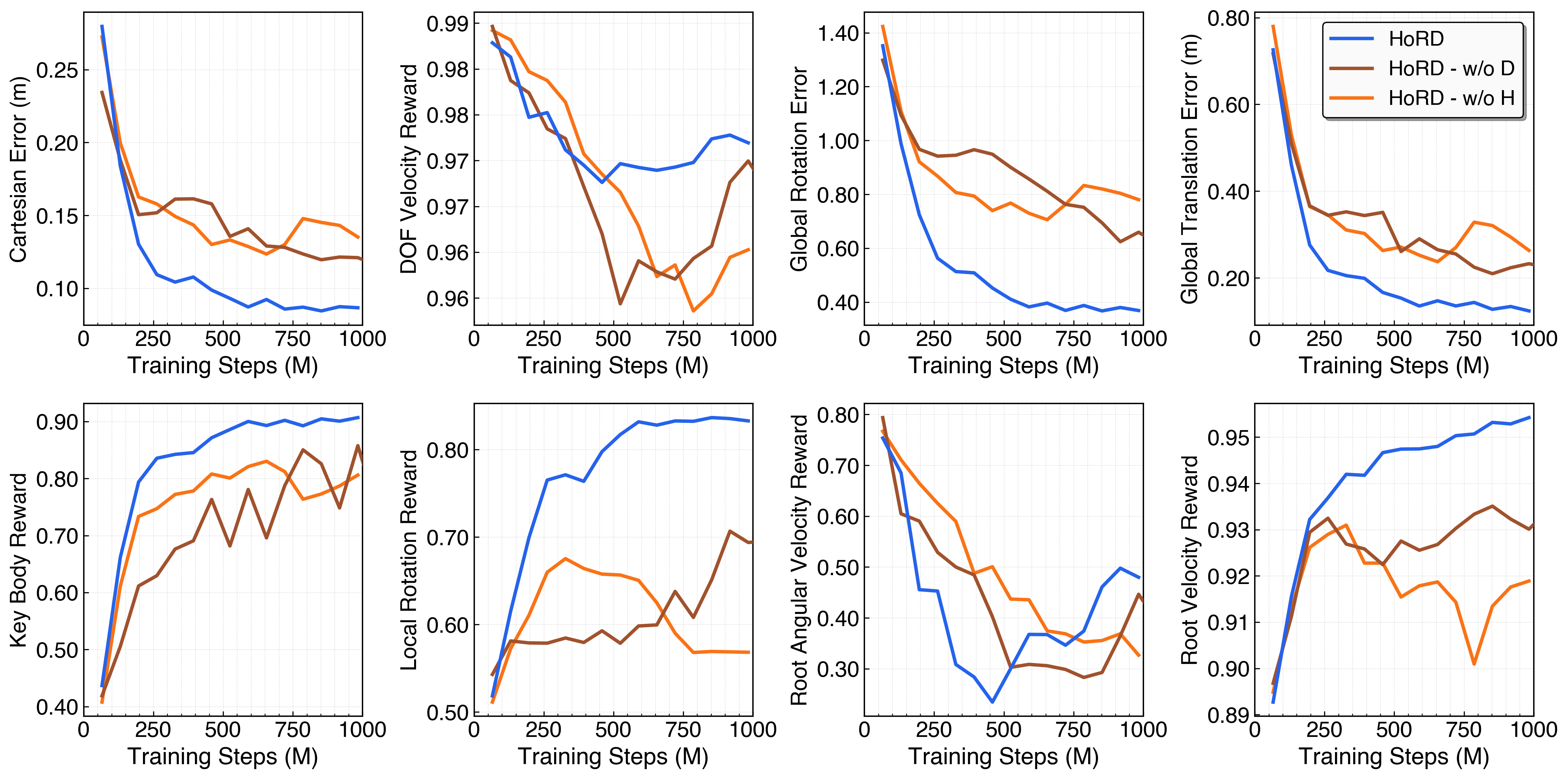}
        \captionsetup{labelformat=empty}
    \end{subfigure}
    \caption{Reward components comparison across different methods. 
        Full HoRD consistently achieves best performance across all reward metrics, 
        demonstrating robust training stability and effectiveness of our approach.}
    \label{fig:training_curves_metrics}
\end{figure}

The expert policy is learned via PPO with dense motion tracking rewards under domain randomization. The student policy is distilled from the expert using online DAgger-style imitation learning, where expert actions are queried during rollouts. Both the teacher and student policies share the same lightweight Transformer encoder backbone with 6 layers, 8 attention heads, model dimension 512, and a $3\times$ feed-forward expansion. HCDR is implemented as a lightweight Q-Former, in which a set of learnable queries attends to a short window of past $(s,a)$ pairs to learn a compact latent summary of interaction history (Sec.~\ref{subsec:HCDR}). This learned history token is appended to the policy input sequence.

The teacher is a goal-conditioned PPO agent learned on dense reference motion under extensive dynamics randomization. The student is distilled from this expert using online DAgger-style imitation learning as described in Sec.~\ref{subsec:framework}. Expert learning uses large-scale parallel simulation with thousands of humanoids and the same terrain configuration (including slopes, stairs, and uneven surfaces), and runs for a total of $10^9$ control steps.

\subsection{Training Resources}
\label{subsec:training_resources}

All methods are trained on the same set of tasks, under the same randomization scheme and environment-step budgets, and all consume SSJR-formatted commands. Unless otherwise stated, experiments are run on a single NVIDIA A100 80\,GB GPU with 32\,GB host RAM using PyTorch~2.3. We evaluate learned policies at $50$\,Hz on held-out motions and terrains from the test split. Success rates and error metrics are averaged over multiple seeds, and significance is assessed using paired $t$-tests with $p{<}0.01$.

\subsection{Training Convergence Curves}
\label{subsec:training_curves}

\begin{wrapfigure}{r}{0.2\textwidth} 
  \centering
  \vspace{-0.5em}
  \includegraphics[width=\linewidth]{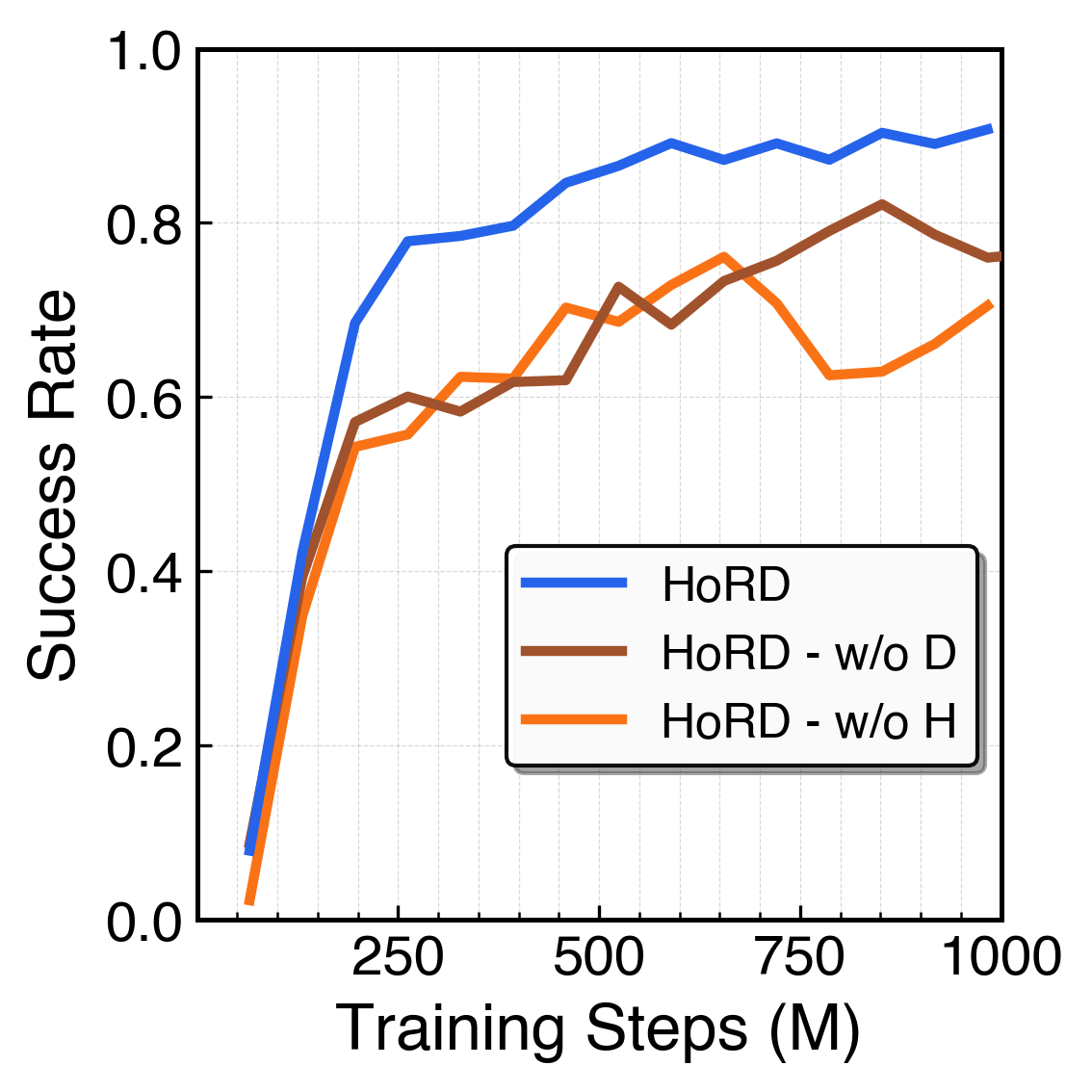}
  \caption{Training convergence curves for different ablation variants, evaluated periodically on the training distribution.}
  \label{fig:training_curves_main}
  \vspace{-2em}
\end{wrapfigure}
We provide additional learning convergence curves for ablation variants to analyze optimization dynamics during training, as shown in Fig.~\ref{fig:training_curves_main} and Fig.~\ref{fig:training_curves_metrics}. 
Policies are trained in IsaacLab with domain randomization (DR) for 1,000M simulation steps. Success rate and intermediate metrics are periodically computed on the training distribution every 500 epochs, corresponding to approximately 65.54M simulation steps per evaluation.
These curves illustrate learning stability, convergence behavior, and optimization dynamics under domain randomization.

\section{Model Details}
\label{app:model}
Both the teacher and student policies adopt a Transformer-based architecture with a 6-layer encoder using 8 attention heads, a model dimension of 512, and a feed-forward dimension of 1,536 (i.e., a $3\times$ expansion). The actor head outputs mean actions and uses a fixed log standard deviation of $-2.9$ with ReLU activation, while the critic is a 4-layer MLP with 1,024 hidden units per layer. The policy takes as input a sequence of observations (5 timesteps for the teacher and 10 timesteps for the student), which is processed by the Transformer encoder. HCDR is implemented as a lightweight Q-Former-style history encoder in which a small set of learnable queries attends to a short window of recent state--action pairs to produce a history token that is appended to the policy input sequence.

\end{document}